%% file: main.tex
\documentclass{article}
\usepackage{iclr2027_conference,times}
\iclrfinalcopy

\newcommand{\nickname}{GRASE}

\usepackage{hyperref}
\hypersetup{hidelinks}
\usepackage{url}
\usepackage{graphicx}
\usepackage{amsmath}
\usepackage{amssymb}

\title{%
\centering
\nickname: Geometry-native Real-to-sim\\[-0.15em]
 Agentic Scene modeling from video%
}

\author{%
\begin{tabular}{c}
\textbf{Zhaorui Lin$^{1}$, Haohan Liu$^{2}$, Zhen Chen$^{3}$, Xiangru Huang$^{4}$\thanks{Corresponding author.}}\\[0.25em]
\textnormal{$^{1}$ Xi'an Jiaotong University (\texttt{xjtu.edu.cn})\quad
$^{2}$ Fuzhou University (\texttt{fzu.edu.cn})}\\
\textnormal{$^{3}$ Adobe Systems (\texttt{adobe.com})\quad
$^{4}$ Westlake University (\texttt{westlake.edu})}
\end{tabular}%
}

\begin{document}

\maketitle
\fancyhead{}

\begin{figure}[h]
    \centering
    \includegraphics[width=\linewidth]{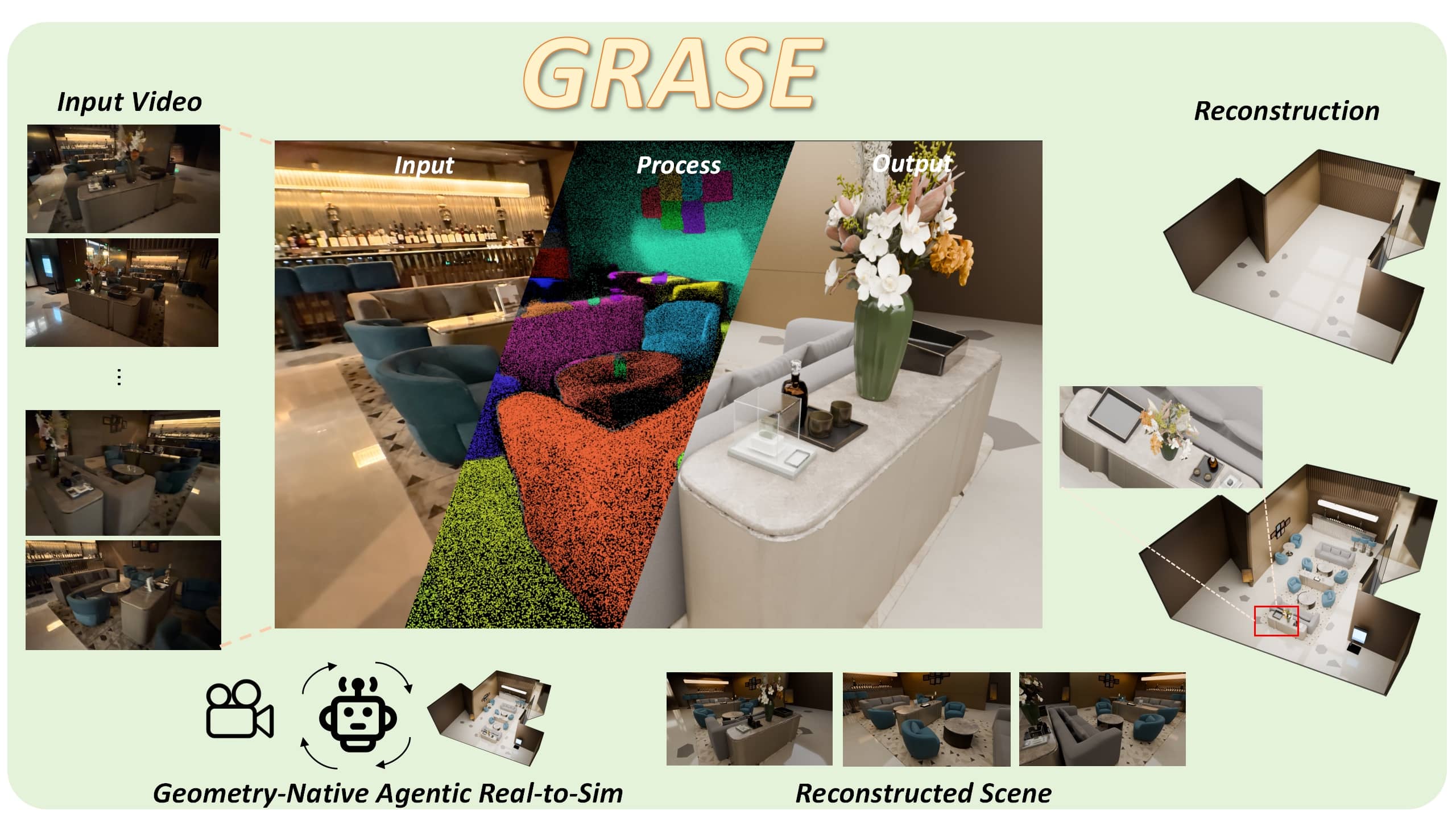}
    \caption{\nickname{} transforms an input video into an editable 3D scene, recovering room structure, object layout, and fine-grained details. The example shows the progression from video observations to reconstruction, together with scene overviews, a close-up, and multiview renderings.}
    \label{fig:teaser}
\end{figure}

\begin{abstract}
Real-to-sim conversion aims to turn a captured video of a real environment into an editable 3D scene while preserving the geometry and layout of what was observed. Existing methods often follow a one-pass, loosely coupled paradigm: each stage receives only limited feedback from previous stages, allowing errors to propagate without reliable correction and leaving the final scene inconsistent. We present GRASE, an agentic framework that closes this loop by treating the captured multi-view video observations as a standing reference throughout scene modeling. Instead of committing to intermediate estimates, GRASE revises its current hypothesis whenever rendered evidence disagrees with the capture, overcoming the fixed intermediate assumptions of conventional pipelines. Specifically, GRASE first fits the architectural shell and its openings from depth and multi-view cues. An agent then places generated assets in a shared coordinate frame while checking their spatial relationships. Finally, each asset's geometry and placement are refined iteratively to improve agreement across the observed views. By coupling these steps through feedback, GRASE localizes structural discrepancies and revises only the responsible components, while preserving cross-view consistency. The resulting 3D scenes are faithful to the input, consistent across views, and suitable for simulations. In our experiments, GRASE substantially outperforms other state-of-the-art methods, including a direct real-to-sim conversion using GPT6-Astra.
\end{abstract}

\input{sections/01_introduction}
\input{sections/02_related}
\input{sections/03_method}

\input{sections/04_experiment}

\input{sections/05_conclusion}
\section*{AI Use Statement}
Generative AI tools were used to aid and polish the writing of this paper, specifically to improve clarity and concision. The authors reviewed all AI-assisted edits and take responsibility for the final content.

\bibliography{iclr2027_conference}
\bibliographystyle{iclr2027_conference}

\appendix

\input{sections/06_appendix}

\end{document}

%% file: sections/01_introduction.tex
\section{Introduction}

Real-to-sim aims to convert real-world observations into simulation-ready 3D environments, providing a scalable way to build training and evaluation data for embodied agents. Video is particularly attractive because it is inexpensive, easy to capture, and naturally records scene appearance and spatial structure. A robust video-based real-to-sim pipeline could therefore greatly reduce the cost of constructing digital twins for robot learning. Real-to-sim requires more than visually plausible reconstruction: the resulting scene must contain accurate geometry, object-level structure, and physically meaningful spatial relationships. Video makes this especially challenging due to occlusions, incomplete viewpoints, motion blur, depth ambiguity, and weakly textured regions. Small geometric errors can further lead to severe simulation failures such as penetration, floating objects, or incorrect contacts.

Existing approaches typically combine scene reconstruction, object segmentation, 3D generation, and scene assembly. However, they often fail to recover large room surfaces, produce inaccurate object geometry from partial observations, or rely on generative models that hallucinate shapes inconsistent with the input video. These errors accumulate during scene composition, resulting in scenes that may look plausible but are geometrically inconsistent and unsuitable for simulation.

An agentic formulation is attractive because real-to-sim naturally requires iterative diagnosis, tool use, and error correction rather than a single feed-forward prediction. In addition, modern VLMs with strong spatial perception capabilities, such as GPT6-Astra, enable a new perceive-then-reconstruct paradigm for real-to-sim conversion. However, we find that applying agentic models such as GPT-6 to recover accurate geometry and topology is not trivial. On the one hand, when using only multi-view images, VLMs struggle to produce accurate 3D geometry at real-world scales. On the other hand, preliminary experiments suggest that even powerful agents such as GPT6-Astra struggle to reason directly over raw point clouds when given estimated point clouds and multi-view camera parameters, frequently making incorrect judgments about geometry, scale, and spatial relationships. This suggests that simply giving 3D observations to an agent is insufficient.

Our key idea is to couple explicit geometric modeling with an agentic verification and repair loop, using the original video as a persistent reference throughout reconstruction. The agent represents the scene through room surfaces and openings, object transforms, and editable object parts, so that discrepancies can be addressed through targeted geometric or topological changes. This leads to two closely connected components: \textbf{room-surface topology and geometry refinement} and \textbf{object-centric layout registration and part-wise geometric refinement}, integrated with \textbf{agentic execution and verification}.

Complex room surfaces, including walls, ceilings, and floors, are traditionally difficult to reconstruct. Using a VLM with strong spatial reasoning capabilities, we propose a perceive-then-reconstruct paradigm that first perceives the architectural topology and then uses geometric primitives and parametric patches to fit regular and irregular surfaces. This paradigm allows us to reliably recover accurate geometry and topology using multi-view depth observations, image boundaries, and agentic spatial reasoning and editing. Notably, this paradigm significantly improves reconstruction results. In the ablation study, we observe a substantial performance drop when this paradigm is not used.

With the room surfaces reconstructed, the agent then aims to reconstruct object-level geometry that aligns with the input video. Agentic models such as GPT6-Astra are good at coding and editing 3D objects using geometric primitives in an agentic loop, but they struggle to match fine and complex real-world geometry. Meshes generated by 3D generation models such as TRELLIS.2, however, are difficult to edit through code. We propose to combine 3D generation models with a part-wise geometric refinement module: we first segment the generated geometry using off-the-shelf segmentation models, then use agentic models to perform part-wise editing, verification, and repair, incorporating geometric cues such as point clouds and image boundaries.

To summarize, our contributions are:
\begin{itemize}
    \item We propose \emph{GRASE}, a geometry-native agentic real-to-sim pipeline that accurately reconstructs the scene from the input video, providing correct and consistent topology recovery for room surfaces.
    \item We propose an object-centric layout registration method that uses agentic part-wise editing to produce refined geometry that better aligns with the input video and is compatible with 3D generation models.
    \item In our experiments, GRASE performs significantly better than its GPT6-Astra counterparts, with or without access to 3D point clouds and 3D generation models, demonstrating the strength of our proposed perceive-then-reconstruct paradigm.
\end{itemize}

%% file: sections/02_related.tex
\vspace{-0.2in}
\section{Related Work}
\label{sec:related}

\vspace{-0.2in}
\noindent\textbf{Image- and Video-Conditioned Compositional Scene Reconstruction.} Reconstructing editable 3D scenes requires recovering object geometry and spatial configuration. Early methods jointly modeled object shapes, poses, layouts, and semantic or instance structure~\citep{nie2020total3dunderstanding,zhang2021holistic,dahnert2021panoptic}. Geometric foundation models subsequently enabled dense correspondence estimation and reconstruction from unposed images~\citep{wang2024dust3r,leroy2024grounding,yang2025fast3r,wang2025vggt}, while spatial memory supports incremental integration of observations~\citep{wang20253d}. Complementing these geometric representations, compositional approaches combine object generation with scene alignment~\citep{shan2026interact3d,zhou2024zero,yao2025cast}. Recent text-to-3D models such as Block3D improve the efficiency of object asset synthesis through block-wise diffusion and confidence-guided token correction~\citep{cui2026block3d}. Other works jointly model multiple objects through diffusion~\citep{huang2025midi,dahnert2024coherent}. Deep Prior Assembly uses completion and candidate alignment, whereas CAST incorporates point-cloud conditioning and physical correction. OVOW~\citep{chen2026one} further extends instance-level mesh reconstruction to dynamic video scenes. Our work focuses on static scene reconstruction from video, combining point-cloud placement constraints with visual feedback to refine object poses and revisit mismatched components. This connects observation-based geometry with iterative correction of an editable, object-level scene representation.

\noindent\textbf{Agentic 3D Scene Generation and Iterative Refinement.} Language-guided systems translate descriptions into layouts, asset selections, and structured scene specifications~\citep{feng2023layoutgpt,yang2024holodeck,wen2023anyhome,ccelen2024design}. Architect and LayoutVLM strengthen spatial guidance through hierarchical image inpainting and differentiable layout optimization, respectively~\citep{wang2024architect,sun2025layoutvlm}. Feedback-based approaches revisit scene states using rendered observations, scene judgment, or reflective tool selection~\citep{hu2024scenecraft,ling2026scenethesis,yang2026sceneweaver,xia2026sage}. Among them, SceneCraft generates and refines Blender programs, while SAGE combines MCP-based orchestration with visual and physical critics. SceneSmith~\citep{pfaff2026scenesmith} organizes hierarchical construction through designer, critic, and orchestrator agents. Spatially contextualized VLMs~\citep{liu2025agentic} further integrate semantic point clouds, scene hypergraphs, and Blender-based verification. These methods motivate combining planning, geometric grounding, and feedback. Our framework centers this process on agreement with an input video: VLM criticism guides Blender-based pose adjustment, followed by comparison and regeneration of mismatched objects. The emphasis is on connecting scene-level feedback to targeted corrections during reconstruction.

\noindent\textbf{Spatial Reasoning and Geometric Grounding in VLMs.} Spatial benchmarks reveal persistent challenges in visual perception, spatial measurement, and reasoning across views or video frames~\citep{fu2024blink,yang2025thinking,zhang2026flatland}. SpatialVLM, SpatialBot, and SpatialRGPT address these challenges through spatial supervision, depth-aware inputs, and region-conditioned reasoning~\citep{chen2024spatialvlm,cai2025spatialbot,cheng2024spatialrgpt}. SPAR~\citep{zhang2026flatland} further scales spatial training data using scenes with 3D ground truth. Complementary methods connect language models to explicit scene representations through point-cloud features, visual prompts, or object embeddings and identifiers~\citep{hong20233d,chen2024ll3da,huang2023chat}. SpatialPIN~\citep{ma2024spatialpin} instead incorporates 3D priors through prompting and tool interaction without additional training. LayoutVLM and spatially contextualized VLMs demonstrate how geometric representations and structured relations support scene modification~\citep{sun2025layoutvlm,liu2025agentic}. Following this principle, we use VLMs for semantic interpretation and corrective feedback, while point-cloud constraints anchor object placement to observed geometry. This couples spatial reasoning with executable pose correction in the reconstructed scene.

%% file: sections/03_method.tex
\section{Geometry-Native Agentic Real-to-Sim}
\label{sec:method}

\begin{figure}[t]
    \centering
    \includegraphics[width=\linewidth]{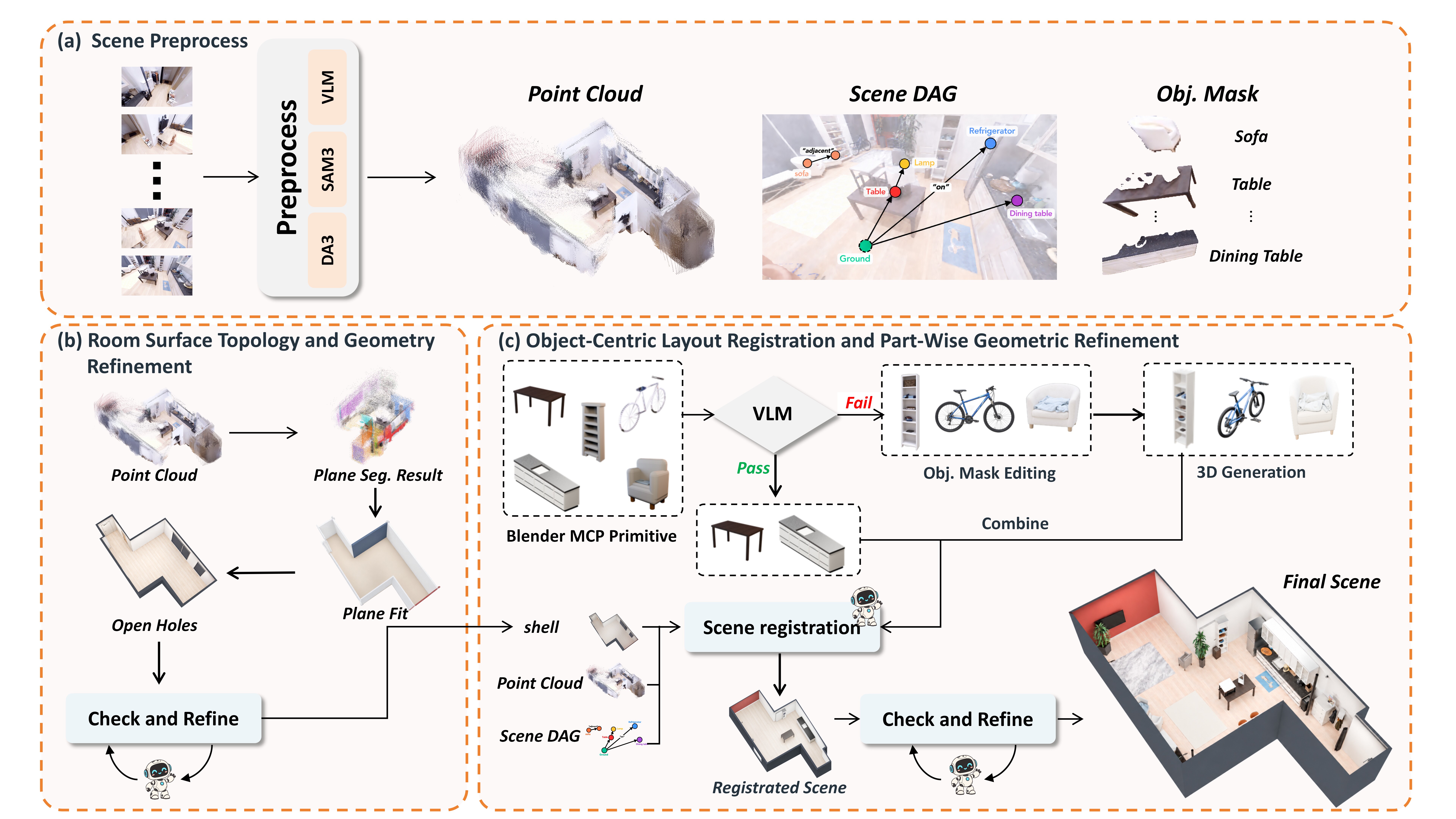}
    \caption{\textbf{Overview of \nickname.} (a) Scene preprocessing extracts a point cloud, a scene DAG, and object masks. (b) Room surface topology and geometry refinement reconstructs the shell and openings. (c) Object-centric layout registration and part-wise geometric refinement align object meshes and correct local shapes. Agentic verification guides both reconstruction stages.}
    \label{fig:grase-pipeline}
\end{figure}

Given a monocular video of a static indoor environment, our goal is to reconstruct an editable 3D scene $\mathcal{S}$ that is consistent with the observed views. The scene consists of room surfaces with explicit openings, one mesh per object with a world-space transform, and the support relations among surfaces and objects.

Figure~\ref{fig:grase-pipeline} summarizes \nickname{}. Instead of letting a model produce geometry directly, \nickname{} exposes room surfaces, opening topology, object transforms, and object parts as variables that are fitted to geometric evidence and edited through constrained operations. After preprocessing, the shell is reconstructed together with its topology (Sec.~\ref{sec:room-surface-refinement}), and objects are generated, registered, and corrected part by part (Sec.~\ref{sec:spatial-refinement}). Appendix~\ref{sec:supp-impl} gives the implementation details, and Table~\ref{tab:supp-hparams} lists all constants.

\paragraph{Scene preprocessing.}
We uniformly sample $T$ frames $\mathcal{I}=\{I_i\}_{i=1}^{T}$ from the video and process them jointly with Depth Anything~3 (DA3), which estimates per-frame depth $z_i$, per-pixel confidence, intrinsics $\mathbf{K}_i$, and camera-to-world poses $\mathbf{T}_i$. Each pixel $\mathbf{u}$ with valid, confident depth is lifted to $\mathbf{T}_i\bigl(z_i(\mathbf{u})\,\mathbf{K}_i^{-1}\tilde{\mathbf{u}}\bigr)$, where $\tilde{\mathbf{u}}$ is its homogeneous coordinate, and merging these points over all frames gives the scene point cloud.

\paragraph{Scene inventory.}
A GPT-6-based VLM organizes the frames into an inventory of surfaces and objects. A wall surface is a connected architectural face whose position and orientation vary continuously, so a smoothly curved wall is one surface, while sharp corners, structural steps, or spatial separation divide surfaces. By matching surfaces across views and inspecting their junctions, the VLM obtains the architectural surfaces $\mathcal{W}$ (walls, floor, and ceiling), their adjacency $\mathcal{A}\subseteq\mathcal{W}\times\mathcal{W}$ (surfaces meeting along a shared physical boundary), and the openings hosted by each surface. For objects, the VLM marks a 2D box around each visible object in every frame and groups the detections by physical identity using appearance, component structure, spatial context, and the recovered cameras. Each group is an \emph{instance}, giving $\mathcal{O}=\{o_j\}_{j=1}^{N}$, where $o_j$ has a category $c_j$ and a 2D box in every frame of $\mathcal{V}_j$, the frames in which it is visible. The VLM further separates large, layout-defining objects (e.g., tables, sofas) from small movable ones (e.g., cups), which are generated and registered differently (Sec.~\ref{sec:spatial-refinement}), and infers a directed acyclic support graph $\mathcal{G}=(\mathcal{W}\cup\mathcal{O},\mathcal{E})$ from visible contact, relative position, and occlusion, where $(u,o_j)\in\mathcal{E}$ means that $u$ supports $o_j$. Finally, SAM~3 segments every surface and object in its observed frames, and lifting the confident depth inside each eroded mask gives point clouds $\mathcal{P}_s$ for surfaces $s\in\mathcal{W}$ and $\mathcal{P}^{\mathrm{obj}}_j$ for objects (Appendix~\ref{sec:supp-preprocess}).

\subsection{Room Surface Topology and Geometry Refinement}
\label{sec:room-surface-refinement}

For each surface, the VLM decides whether a plane suffices or whether connected parametric patches are needed. Throughout, $\rho(x)=2\bigl(\sqrt{1+x^{2}}-1\bigr)$ denotes the soft-$L_1$ penalty, which limits the influence of outliers.

\paragraph{Regular surfaces.}
A planar surface $s$ is represented as $\Pi_s=\{\mathbf{x}:\mathbf{n}_s^{\top}\mathbf{x}+d_s=0\}$ with $\|\mathbf{n}_s\|_2=1$, initialized by RANSAC on $\mathcal{P}_s$ and refit to its inliers by SVD. The adjacency fixes the topology of the shell. If each wall were then cropped to its own points, neighboring walls would rarely meet exactly, leaving gaps or overlaps at the corners. We instead take a wall's extent from its neighbors. Three mutually adjacent surfaces $s,t,r$ in $\mathcal{A}$, such as two walls and the floor, meet at the corner $\mathbf{v}_{str}$ that solves
\begin{align}
[\mathbf{n}_s\;\mathbf{n}_t\;\mathbf{n}_r]^{\top}\mathbf{v}_{str}=-[d_s\;d_t\;d_r]^{\top},
\end{align}

and these corners bound each wall as a finite polygon. Adjacent surfaces reference the same corners, which are recomputed whenever an incident plane moves. The adjacency therefore fixes the topology of the shell while the planes remain free to change: refining or editing one plane moves the shared corners with it and never opens a gap at a junction.

The initial fits are accurate in the middle of each surface but not necessarily at its junctions, the lines where it meets its neighbors, such as the corner between two walls or the base of a wall. To refine the planes, we fit the junctions observed in the images: notice that a junction is typically visible in multiple frames and each sighting of junction $k$ in frame $i$ is one observation $(i,k)\in\mathcal{J}$. For this observation, $B_{ik}$ contains pixels sampled along the junction in frame $i$, away from occluders and image borders, and $\boldsymbol{\ell}_{ik}(\boldsymbol{\theta})$ is where the current planes place the junction in that frame: the projection of the two planes' intersection line, scaled so that $\boldsymbol{\ell}_{ik}^{\top}\tilde{\mathbf{u}}$ is a signed pixel distance. All planes are refined jointly by
\begin{equation}
    \begin{aligned}
        \min_{\boldsymbol{\theta}}\;
        &\sum_{s}\frac{1}{|\mathcal{P}_s|}\sum_{\mathbf{p}\in\mathcal{P}_s}
        \rho\!\left(\frac{\mathbf{n}_s^{\top}\mathbf{p}+d_s}{\sigma_s}\right) +\frac{\lambda_{\mathrm{jct}}}{|\mathcal{J}|}\sum_{(i,k)\in\mathcal{J}}\frac{1}{|B_{ik}|}\sum_{\mathbf{u}\in B_{ik}}
        \rho\!\left(\frac{\boldsymbol{\ell}_{ik}(\boldsymbol{\theta})^{\top}\tilde{\mathbf{u}}}{\sigma_{\mathrm{px}}}\right)\\
        &+\lambda_{\mathrm{prior}}\sum_{m}\Bigl(\frac{\theta_m-\theta_{m,0}}{\epsilon_m}\Bigr)^{2},
        \quad |\theta_m-\theta_{m,0}|\le\epsilon_m,
    \end{aligned}
\label{eq:shell-joint-fit}
\end{equation}
where $s$ ranges over the planar surfaces, $\boldsymbol{\theta}$ collects the plane offsets and normal corrections, $\sigma_s$ and $\sigma_{\mathrm{px}}$ are fixed 3D and pixel scales, $\lambda_{\mathrm{jct}}$ and $\lambda_{\mathrm{prior}}$ are fixed weights, and $\epsilon_m$ bounds the change of each parameter from its initial value $\theta_{m,0}$. We solve this bounded robust least-squares problem with the cameras and $\mathcal{A}$ fixed (Appendix~\ref{sec:supp-shell}). Because the shared corners are recomputed from the planes at every evaluation, the shell stays closed throughout the optimization. The resulting polygons are finally triangulated into editable Blender meshes.

\paragraph{Irregular surfaces.}
A surface that no single plane describes, such as a curved wall, is represented by connected patches, each mapping local coordinates $(\xi,\eta)$ to a 3D point,
\begin{equation}
\mathbf{f}(\xi,\eta)=\mathbf{C}\,\mathbf{b}(\xi,\eta),
\label{eq:irregular-surface-parameterization}
\end{equation}
where $\mathbf{b}$ contains fixed basis functions and $\mathbf{C}$ their 3D coefficients. We use two families of patches. A \emph{periodic loft} covers a region that wraps around the room; its coordinates are the horizontal angle and the height, and its horizontal position combines periodic trigonometric terms with quadratic terms in height. A \emph{local quadratic displacement field} offsets a reference plane along its normal by a quadratic function, and a vertical reference normal makes it a height field. With the reference frame fixed, both families are linear in $\mathbf{C}$ and are fitted by least squares, with an envelope prior for poorly observed parts of a loft and stronger penalties on the quadratic terms of a displacement field to avoid unsupported curvature. The VLM specifies how neighboring patches meet. At a sharp junction, we intersect the two surfaces and trim both parameter domains by the intersection curve, so the seam vertices coincide after triangulation. Where the transition is smooth, we blend the gradient-normalized implicit functions of the two patches with a smooth maximum and project nearby vertices onto the blend, whose width falls to zero near boundaries that must stay fixed. Please refer Appendix~\ref{sec:supp-shell} for details.

\input{figures/tex/room_surface_refinement}

\paragraph{Openings and inferred geometry.}
Doors, windows, recesses, and passages are modeled as geometry rather than as appearance changes on an intact wall. For each inventoried opening, the VLM traces its visible rim in the source frames and determines whether it leads to a rear surface or an open passage. Rays through the traced rim are intersected with the host surface to initialize a 3D rim. We cut the enclosed region from the surface and build side surfaces along the rim, together with the rear surface when it is observed. Openings visible on a rear surface are built on that surface in turn, which handles nested structures. Occluded parts of the shell, such as the wall behind a curtain, are completed by extending the fitted surfaces under their parameterizations up to their junctions, and are marked as inferred rather than observed. The finished shell is rendered from the source cameras, and remaining discrepancies in surface shape, junctions, or opening rims go to the verification loop. Figure~\ref{fig:room-surface-refinement-workflow} illustrates this workflow.

\subsection{Object-Centric Layout Registration and Part-Wise Geometric Refinement}
\label{sec:spatial-refinement}

With the shell in place, we build a mesh $M_j$ and a world-space transform for every object $o_j$. The support graph $\mathcal{G}$ sets the order: large objects come first, and an object is placed only once everything supporting it is stable. Actual contact is checked against the supporting meshes during registration. 

\paragraph{Object initialization}
Large objects, such as tables, chairs, and other furniture, are built directly in Blender by the VLM through the Model Context Protocol (MCP), using simple primitives for their major components. This gives clean, editable geometry whose parts follow the object's structure. When the VLM judges the resulting shape too poor, the object is generated with TRELLIS.2 instead, and small objects are always generated with TRELLIS.2 (Figure~\ref{fig:object-generation-placement}). The VLM builds each primitive model in place, using the location and extent of the object's point cloud $\mathcal{P}^{\mathrm{obj}}_j$, whereas a TRELLIS.2 mesh arrives in its own canonical frame and is registered to $\mathcal{P}^{\mathrm{obj}}_j$ by ICP. Since the point cloud covers only the visible side of the object and is noisy, either alignment can still be off in size or pose. The agent therefore selects three frames that together view the object from as many sides as possible, and the VLM compares renderings of the placed object with these frames and adjusts its size and pose accordingly. A detailed pipeline can be found in the Appendix~\ref{sec:supp-registration}.

\input{figures/tex/object_generation_placement}

\paragraph{Part-wise geometric refinement.}
A generated mesh may merge several components into one piece, such as a sofa and its cushions, and a single transform cannot fit them all: once the sofa body matches the observations, the cushions may still be misplaced. The VLM finds such a component by comparing renderings with the source frames, and the agent selects its region on the mesh by casting rays through the component's mask in several renderings. The component is then re-aligned to its own points while the rest of the mesh stays fixed. A component with wrong proportions, such as the sofa backrest in Figure~\ref{fig:part-stretch}, is stretched or contracted along a local axis $\mathbf{a}$ between a fixed anchor $\mathbf{x}_{\mathrm{fix}}$ and a moving anchor $\mathbf{x}_{\mathrm{mov}}$:
\begin{equation}
\mathbf{v}'=\mathbf{v}+S\bigl(\tau(\mathbf{v})\bigr)\,\Delta\,\mathbf{a},\qquad
\tau(\mathbf{v})=\frac{(\mathbf{v}-\mathbf{x}_{\mathrm{fix}})^{\top}\mathbf{a}}{(\mathbf{x}_{\mathrm{mov}}-\mathbf{x}_{\mathrm{fix}})^{\top}\mathbf{a}},
\label{eq:part-stretch}
\end{equation}
for every vertex $\mathbf{v}$ of the component, where $\Delta$ is the signed length correction and $S(\tau)=3\tau^{2}-2\tau^{3}$, with $\tau$ clamped to $[0,1]$, is the cubic smoothstep, so the edit fades smoothly into the fixed geometry. An edit is kept only if the component matches the observations better without disturbing the rest of the object (Appendix~\ref{sec:supp-part}).

\input{figures/tex/part_stretch}

\paragraph{Cross-object refinement.}
The previous steps place each object on its own, so relations between objects are not yet enforced: a computer may float above the desk it stands on, or two neighboring objects may interpenetrate. We therefore check every support relation in $\mathcal{G}$ against the actual geometry. An object is moved along the support normal until its base touches the supporting mesh, and samples across the contact region are checked for floating or penetration; wall-mounted and suspended objects are held to their observed attachment regions instead. Neighboring objects are also checked for interpenetration, and each correction is verified against the source frames through agent. Because contact is defined relative to the supporting mesh, objects are processed in the order of $\mathcal{G}$, and a later change to a support triggers new checks of the objects it supports (Appendix~\ref{sec:supp-cross}).

%% file: figures/tex/room_surface_refinement.tex
\begin{figure}[!htbp]
    \centering
    \includegraphics[width=\linewidth]{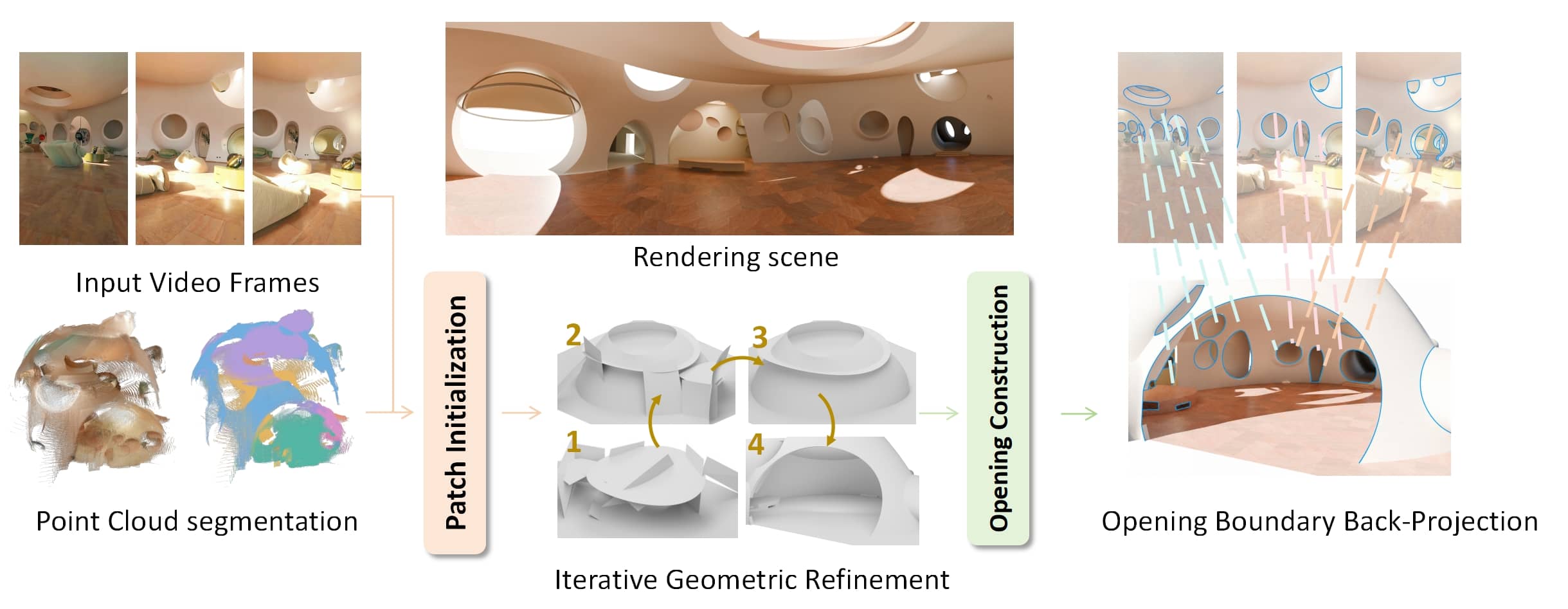}
    \caption{\textbf{Room-surface reconstruction and refinement.} Input video frames and the segmented point cloud guide patch initialization. The patches are iteratively refined using geometric evidence and comparisons between source frames and renderings. Observed opening boundaries are back-projected onto the fitted surfaces to construct explicit openings.}
    \label{fig:room-surface-refinement-workflow}
\end{figure}

%% file: figures/tex/object_generation_placement.tex
\begin{figure}[t]
    \centering
    \includegraphics[width=\linewidth]{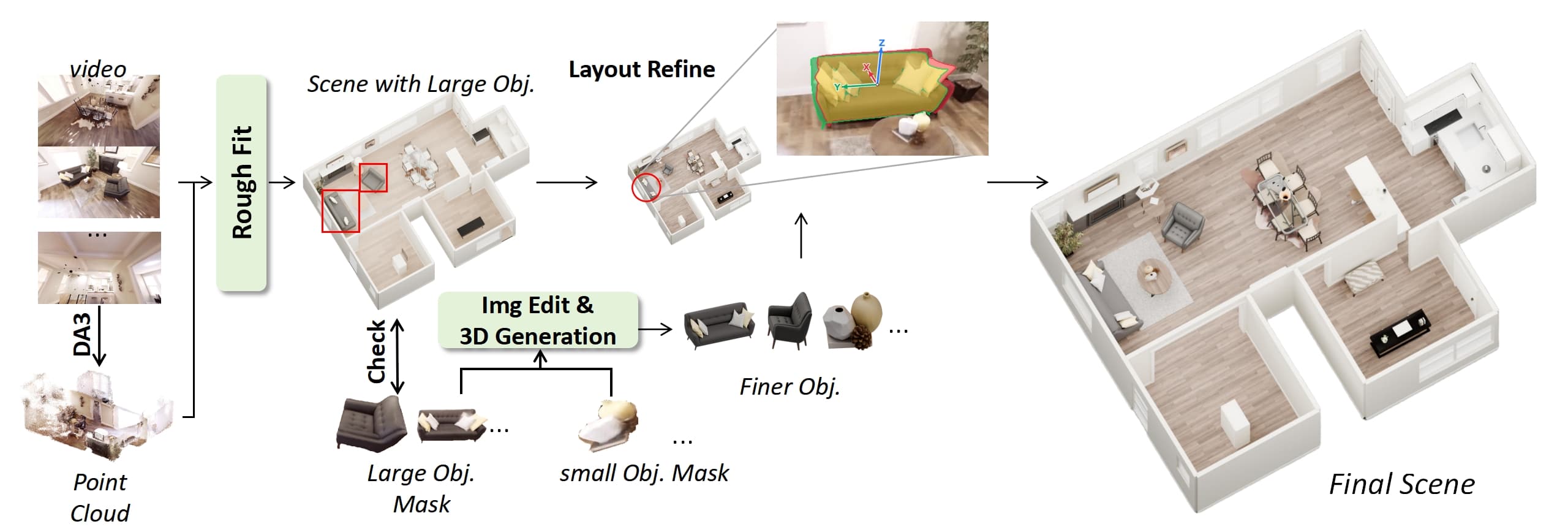}
    \caption{\textbf{Object generation and placement.} The large objects in ``Scene with Large Obj.'' are built from primitives through Blender MCP. The VLM checks them against source views; only models with incorrect shapes or structures are replaced using TRELLIS.2. Small objects are generated directly with TRELLIS.2. All objects are then registered and refined to form the final scene.}
    \label{fig:object-generation-placement}
\end{figure}

%% file: figures/tex/part_stretch.tex
\begin{figure}[!htbp]
    \centering
    \includegraphics[width=\linewidth,trim=0 250bp 0 291.667bp,clip]{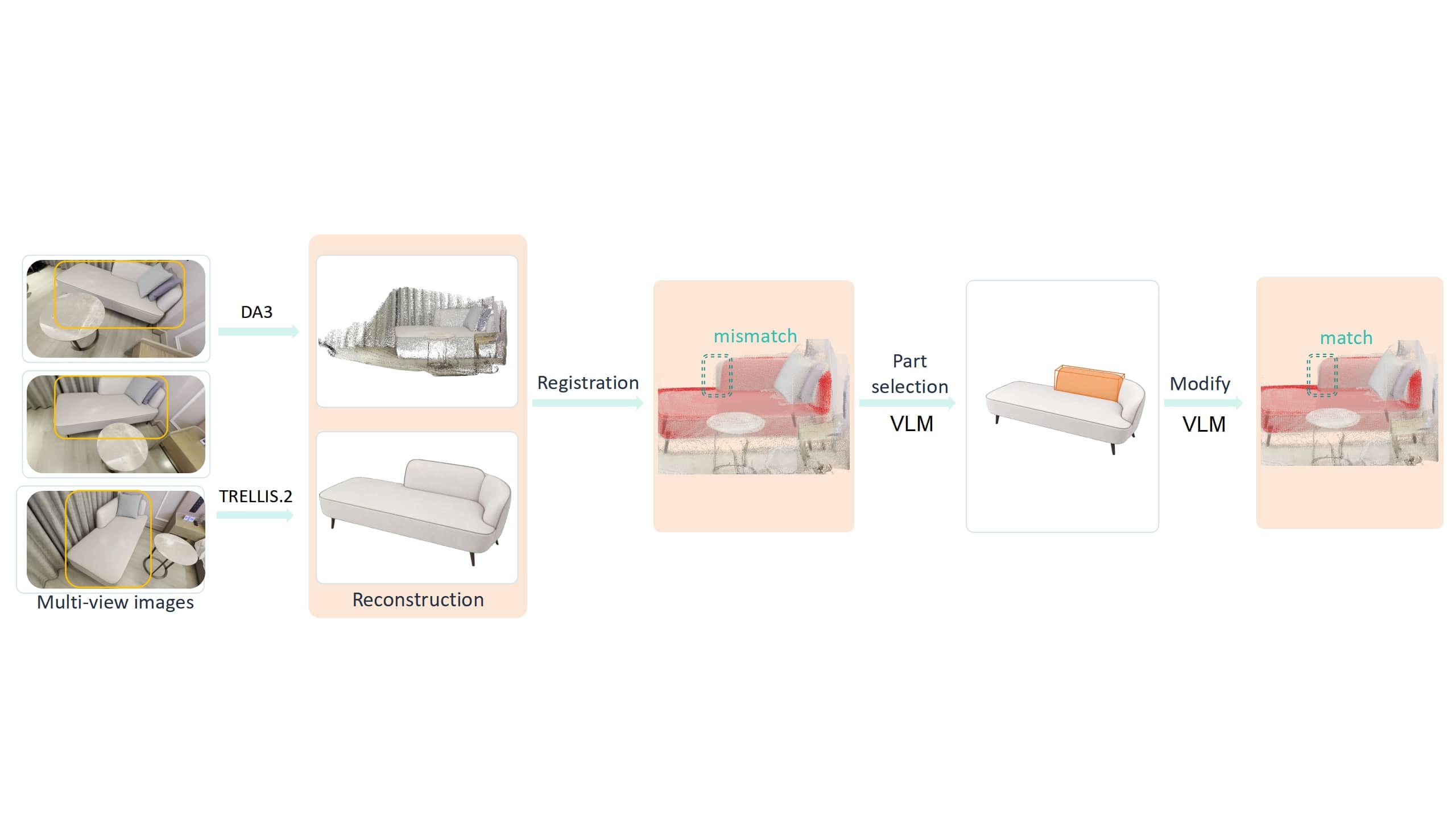}
    \caption{\textbf{Part-wise geometric refinement of a large object.} After object registration, a local mismatch remains in the sofa backrest. The VLM identifies the affected part and guides a constrained stretch that improves its alignment with the observed geometry while preserving the rest of the object.}
    \label{fig:part-stretch}
\end{figure}

%% file: sections/04_experiment.tex
\section{Experiments}
\label{sec:experiments}
\suppressfloats[t]
We evaluate GRASE at the architectural, object, and scene levels, and examine the contribution of geometric feedback through ablation. Input-view rendering provides a complementary measure of agreement with the captured observations.

\subsection{Experimental Setup}

\paragraph{Datasets and baselines.}
We compare GRASE against three baselines on the same eight Replica and fifteen Imaginarium scenes: \textbf{SAM3D}; \textbf{GPT-6 vanilla}, which constructs scenes directly from visual observations; and \textbf{GPT-6 + TRELLIS.2 + point cloud}, which additionally has access to point-cloud/camera information and TRELLIS.2 asset generation. Generation inputs and alignment protocols differ across methods.

\paragraph{Metrics.}
Both datasets share metric definitions: Chamfer Distance (CD), normal consistency (NC), and F-score at 5\,cm (F@5cm) for geometry; instance mIoU for reconstructed--reference object agreement; and PSNR, SSIM, and LPIPS for input-view renderings against reference images. Object-level metrics use the full ground-truth (GT) object count as the denominator. Architectural shells are evaluated separately from scene geometry. Further evaluation details are provided in Appendix~\ref{sec:supp-evaluation}.

\begin{table*}[t]
  \centering
  \caption{Object- and scene-level reconstruction and input-view rendering on Imaginarium and Replica. PC denotes point cloud; -- indicates unavailable rendering metrics.}
  \label{tab:main_results}
  \label{tab:imaginarium_geometry}
  \label{tab:imaginarium_rendering}
  \label{tab:replica_main_offices}
  \label{tab:benchmark_rendering}
  \label{tab:replica_rendering_ablation}
  \setlength{\tabcolsep}{4pt}
  \resizebox{\textwidth}{!}{%
  \begin{tabular}{@{}lcccccccccc@{}}
    \toprule
    & \multicolumn{4}{c}{Object-level} & \multicolumn{3}{c}{Scene-level} & \multicolumn{3}{c}{Input-view rendering} \\
    \cmidrule(lr){2-5}\cmidrule(lr){6-8}\cmidrule(l){9-11}
    Method & mIoU$\uparrow$ & NC$\uparrow$ & CD (cm)$\downarrow$ & F@5cm$\uparrow$
           & CD (cm)$\downarrow$ & NC$\uparrow$ & F@5cm$\uparrow$
           & PSNR$\uparrow$ & SSIM$\uparrow$ & LPIPS$\downarrow$ \\
    \midrule
    \multicolumn{11}{@{}l}{\textit{Imaginarium}} \\
    SAM3D & 0.0482 & 0.6182 & 7.1482 & 0.0648 & 72.1585 & 0.5892 & 0.1006 & -- & -- & -- \\
    GPT-6 vanilla & 0.1066 & 0.6490 & 18.4091 & 0.1144 & 18.1588 & 0.6844 & 0.3449 & 11.8223 & 0.5342 & 0.6481 \\
    GPT-6 + TRELLIS.2 + PC & 0.2066 & 0.6850 & 11.8184 & 0.2062 & 10.9977 & 0.7946 & 0.5615 & 12.5762 & 0.5994 & 0.5606 \\
    GRASE (Ours) & \textbf{0.7425} & \textbf{0.7929} & \textbf{2.1936} & \textbf{0.8536} & \textbf{3.1412} & \textbf{0.9044} & \textbf{0.8866} & \textbf{15.5692} & \textbf{0.6667} & \textbf{0.3381} \\
    \midrule
    \multicolumn{11}{@{}l}{\textit{Replica}} \\
    SAM3D & 0.1566 & 0.6666 & 4.9177 & 0.1339 & 63.9258 & 0.6050 & 0.1908 & -- & -- & -- \\
    GPT-6 vanilla & 0.1897 & 0.6346 & 11.9005 & 0.1742 & 16.3887 & 0.6177 & 0.3452 & 11.1367 & 0.6943 & 0.5488 \\
    GPT-6 + TRELLIS.2 + PC & 0.1325 & 0.5670 & 11.0108 & 0.1063 & 19.9107 & 0.6129 & 0.2165 & 10.8904 & 0.6373 & 0.5785 \\
    GRASE w/o geo.\ critic & 0.1577 & 0.7076 & 9.7786 & 0.1740 & 9.9572 & 0.7729 & 0.5418 & 14.5470 & 0.6965 & 0.4190 \\
    GRASE (Ours) & \textbf{0.4121} & \textbf{0.7271} & \textbf{4.5207} & \textbf{0.4507} & \textbf{5.7963} & \textbf{0.8287} & \textbf{0.6477} & \textbf{16.3674} & \textbf{0.7454} & \textbf{0.3595} \\
    \bottomrule
  \end{tabular}%
  }
\end{table*}

\begin{table}[t]
  \centering
  \caption{Architectural shell reconstruction: (a) Imaginarium comparison; (b) seven-scene critic ablation. See Appendix~\ref{sec:supp-evaluation} for the ablation protocol and units.}
  \label{tab:imaginarium_shell_geometry}
  \label{tab:shell_ablation}
  \small
  \begin{minipage}[b]{0.56\linewidth}
    \centering
    \setlength{\tabcolsep}{3pt}
    \resizebox{\linewidth}{!}{%
    \begin{tabular}{@{}lccc@{}}
      \multicolumn{4}{c}{(a) Imaginarium} \\
      \toprule
      Method & CD (cm)$\downarrow$ & NC$\uparrow$ & F@5cm$\uparrow$ \\
      \midrule
      GPT-6 vanilla & 29.6426 & 0.7916 & 0.3474 \\
      GPT-6 + TRELLIS.2 + PC & 6.5474 & 0.8785 & 0.6729 \\
      GRASE (Ours) & \textbf{2.9734} & \textbf{0.9548} & \textbf{0.8263} \\
      \bottomrule
    \end{tabular}
    }
  \end{minipage}\hfill
  \begin{minipage}[b]{0.40\linewidth}
    \centering
    \setlength{\tabcolsep}{3pt}
    \resizebox{\linewidth}{!}{%
    \begin{tabular}{@{}lccc@{}}
      \multicolumn{4}{c}{(b) Critic ablation} \\
      \toprule
      Method & CD$\downarrow$ & NC$\uparrow$ & F-score$\uparrow$ \\
      \midrule
      w/o Critic & 55.74 & 0.859 & 0.134 \\
      Ours       & \textbf{3.33} & \textbf{0.965} & \textbf{0.767} \\
      \bottomrule
    \end{tabular}
    }
  \end{minipage}
\end{table}

\subsection{Quantitative Comparison}

\paragraph{Object and scene reconstruction.}
GRASE improves reconstruction at both the object and scene levels on both datasets (Table~\ref{tab:main_results}). Relative to the strongest baseline in scene-level CD on each dataset, it reduces the error from 11.00 to 3.14\,cm on Imaginarium and from 16.39 to 5.80\,cm on Replica, corresponding to reductions of 71.4\% and 64.6\%, respectively. The improvements extend to object agreement, with mIoU reaching 0.7425 and 0.4121. GRASE also achieves the best input-view rendering scores among the methods with reported rendering metrics, indicating that the geometric gains are accompanied by closer agreement with the input images.

\paragraph{Access to assets and point clouds.}
The two GPT-6 baselines exhibit different trends across datasets. Adding TRELLIS.2 and point-cloud access improves Imaginarium scene-level CD from 18.16 to 11.00\,cm, but increases the error on Replica from 16.39 to 19.91\,cm; the rendering metrics follow the same trend. In contrast, GRASE improves reconstruction on both datasets. This comparison suggests that access to generative assets and geometric observations alone does not ensure accurate scene modeling.

\paragraph{Architectural shell reconstruction.}
GRASE also improves the geometry of the room shell (Table~\ref{tab:imaginarium_shell_geometry}a). On Imaginarium, it reduces CD from 6.55\,cm for GPT-6 + TRELLIS.2 + point cloud to 2.97\,cm, a 54.6\% reduction. Its NC of 0.955 and F@5cm of 0.826 further indicate improved surface orientation and geometric agreement. These results complement the object- and scene-level evaluation by assessing the architectural structure separately.

\subsection{Effect of Geometric Feedback}

We ablate the joint point-cloud and hard-crease critic while retaining the remaining agent workflow. On Replica (Table~\ref{tab:main_results}), enabling the critic reduces object-level CD from 9.78 to 4.52\,cm and scene-level CD from 9.96 to 5.80\,cm, reductions of 53.8\% and 41.8\%, respectively. Object mIoU increases from 0.1577 to 0.4121, accompanied by an increase in input-view PSNR from 14.55 to 16.37\,dB. Thus, the contribution of geometric feedback is reflected in both object- and scene-level reconstruction, as well as input-view rendering.

The separate seven-scene shell ablation (Table~\ref{tab:shell_ablation}b) shows the same direction of improvement: CD decreases from 55.74 to 3.33, a 94.0\% reduction, while NC increases from 0.859 to 0.965 and F-score from 0.134 to 0.767. These shell results use the alignment and source-coordinate units specified in Appendix~\ref{sec:supp-evaluation}. Together, the ablations support the role of geometric feedback in refining both architectural surfaces and the assembled scene.

\subsection{Qualitative Analysis}

Figure~\ref{fig:qualitative_four_scene_comparison} compares four scenes using matched camera poses and intrinsics for the input and the first four rendered columns. These views expose differences in room boundaries, furniture proportions, and relative object placement; the last column shows overviews of the GRASE reconstructions. Additional viewpoints are provided in Figure~\ref{fig:qualitative_full_multiview}.

The component-level examples illustrate how the editable representation supports targeted correction. Figure~\ref{fig:room-surface-refinement-workflow} shows the reconstruction of curved room surfaces and explicit openings. In Figure~\ref{fig:part-stretch}, a local backrest mismatch remains after the sofa is registered; editing that part improves its alignment while preserving the rest of the asset. These examples illustrate corrections at the architectural and part levels that complement the aggregate reconstruction metrics.

\input{comparison_multiview/four_scene_multiview}

%% file: comparison_multiview/four_scene_multiview.tex
\begin{figure*}[t]
\centering
\includegraphics[width=\textwidth]{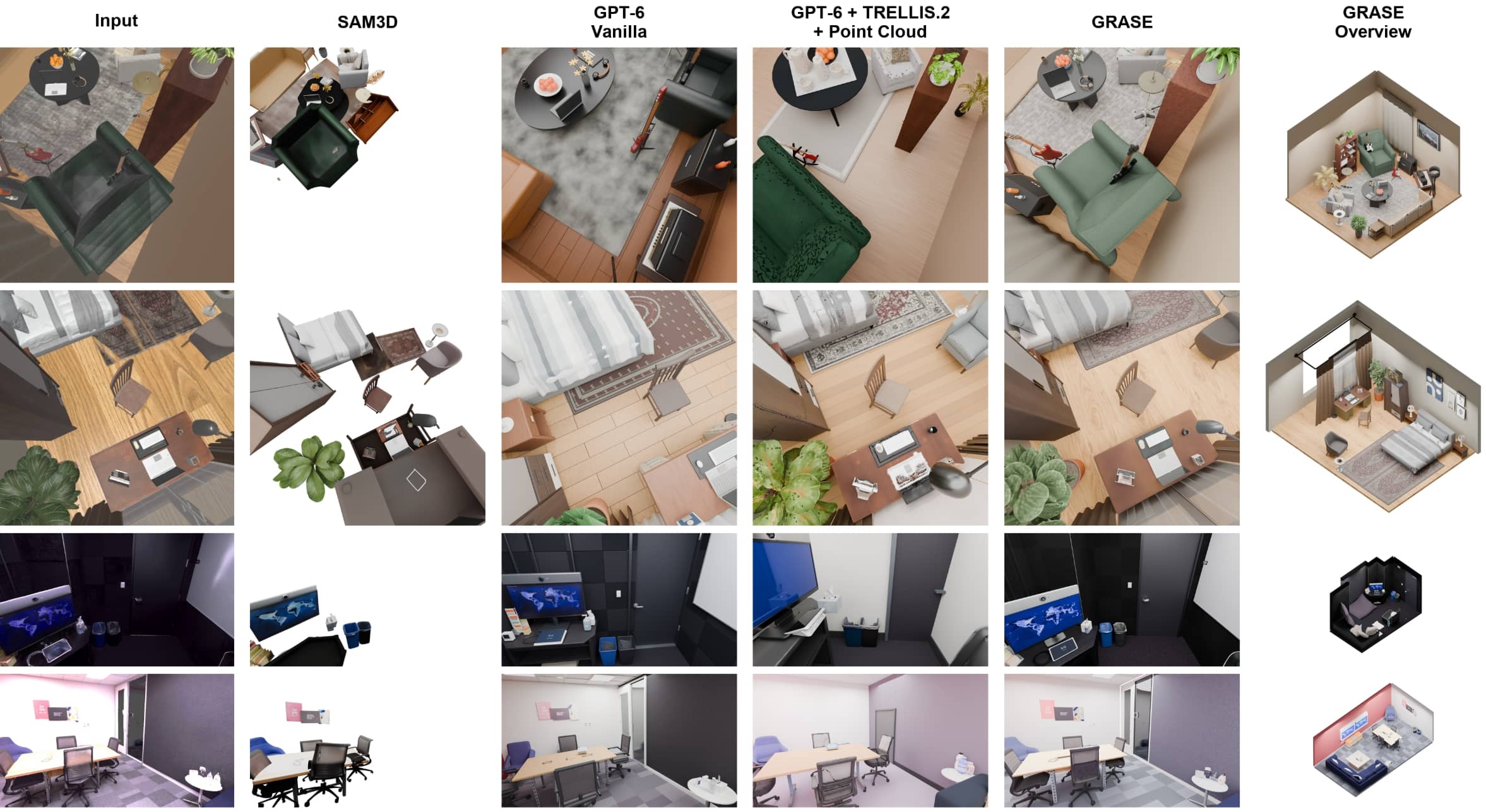}
\caption{\textbf{Qualitative comparison on four scenes.} Each row shows the input, matched-camera renderings from SAM3D, GPT-6 vanilla, GPT-6 + TRELLIS.2 + point cloud, and GRASE, followed by a novel-view overview of the GRASE scene. Figure~\ref{fig:qualitative_full_multiview} provides additional input-camera viewpoints for the same scenes.}
\label{fig:qualitative_four_scene_comparison}
\label{fig:qualitative_office_multiview}
\end{figure*}

%% file: sections/05_conclusion.tex
\section{Conclusion}
\label{sec:limitations}
We present GRASE, a geometry-native agentic framework that reconstructs editable indoor scenes from monocular videos through shell modeling, object layout registration, and part-wise refinement under multi-view verification, outperforming prior methods on Replica and Imaginarium. GRASE has several limitations. First, reconstruction remains limited in heavily occluded regions where the input video provides insufficient geometric evidence; such regions are completed by extending fitted surfaces or by generative priors and are marked as inferred rather than observed. Second, the scene inventory, support graph, and verification decisions rely on VLM judgments, so an early error, such as a missed junction or an incorrect support relation, can persist when the rendered evidence is ambiguous. Third, GRASE assumes a static scene and reconstructs geometry and layout only; articulation, materials, and physical properties required by many simulators are not yet recovered. Future work will extend the framework to articulated and dynamic objects, outdoor scenes, estimate physical and material properties from video, and replace part of the VLM-based verification with learned geometric critics to reduce cost and improve robustness.

%% file: sections/06_appendix.tex
\input{appendix_techinical_details/technical_details.tex}
\input{comparison_multiview/four_scene_multiview_full}

\input{appendix_qualitative/tex/gallery}

%% file: appendix_techinical_details/technical_details.tex
\section{Implementation Details of \nickname{}}
\label{sec:supp-impl}

This section gives the details omitted from Sec.~\ref{sec:method}. We keep the notation of the main paper, and Table~\ref{tab:supp-hparams} lists all constants.

\subsection{Scene Preprocessing and Inventory}
\label{sec:supp-preprocess}

\paragraph{Depth and cameras.}
We sample $T$ frames uniformly from each video and run DA3 on all of them jointly, obtaining depth maps, per-pixel confidence, intrinsics, and camera poses. Before back-projection, we check the camera and pixel-coordinate conventions of the DA3 outputs, discard invalid depth, and filter low-confidence pixels. A retained pixel $\mathbf{u}=(u,v)$ of frame $i$ is back-projected as
\begin{equation}
\mathbf{x}_w=\mathbf{T}_i\bigl(z_i(\mathbf{u})\,\mathbf{K}_i^{-1}\tilde{\mathbf{u}}\bigr),\qquad \tilde{\mathbf{u}}=[u,v,1]^{\top},
\label{eq:supp-backprojection}
\end{equation}
where $z_i(\mathbf{u})$ is the camera-$Z$ depth, $\mathbf{K}_i$ the intrinsic matrix, and $\mathbf{T}_i$ the rigid transformation from the camera frame of $I_i$ to the world frame.

\paragraph{Segmented point clouds.}
For objects, each SAM~3 mask is eroded with a kernel of size $r_{\mathrm{erode}}$. Inside the eroded mask, we keep valid, positive-depth pixels whose DA3 confidence exceeds $\max(\tau_{\mathrm{conf}},Q_{\kappa})$, where $\tau_{\mathrm{conf}}$ is a fixed floor and $Q_{\kappa}$ is the $\kappa$-quantile of the confidences inside that mask. The retained pixels are subsampled to at most $N_{\mathrm{view}}$ per view and back-projected with Eq.~\eqref{eq:supp-backprojection}. For surfaces, we likewise discard unreliable depth and points near segmentation boundaries, and we aggregate the points of each physical surface across frames using the cross-view correspondences in the inventory.

\paragraph{Scene inventory.}
In every sampled frame, the VLM places a 2D bounding box around the visible extent of each wall surface. The boxes localize surfaces for inspection only; they do not define boundaries or geometry. The VLM compares structural cues inside and around the boxes across successive and complementary frames to decide which observations show the same physical wall, accounting for viewpoint changes and partial occlusion, and it examines shared junctions across views to decide which surfaces are adjacent. Each visible opening is associated with its host surface. The surface inventory thus records cross-view correspondences, per-frame boxes and visibility, adjacency, and opening-to-surface relations.

Objects are handled in the same way. The VLM assigns a semantic category and a 2D box to every visible object and associates detections across views by consistency in appearance, component structure, and spatial context, using the recovered cameras to account for viewpoint changes. Repeated observations of one object share an identifier, while distinct objects of the same category keep separate identifiers. Large, layout-defining objects are distinguished from small ones by their observed extent and role in the scene, and support is inferred from visible contact, relative position, and occlusion cues across views.

\subsection{Room Surface Reconstruction}
\label{sec:supp-shell}

\paragraph{Plane initialization.}
For each planar surface, we discard unreliable depth samples and points near segmentation boundaries. Every nondegenerate three-point sample of $\mathcal{P}_s$ defines a candidate plane, and RANSAC keeps the candidate supported by the most points within distance $\tau_{\mathrm{ransac}}$. We then refit the inliers by SVD of their centered coordinates, taking the right singular vector with the smallest singular value as $\mathbf{n}_s$ and setting $d_s=-\mathbf{n}_s^{\top}\bar{\mathbf{p}}$, where $\bar{\mathbf{p}}$ is the inlier centroid.

\paragraph{Corners and polygons.}
Every triple of mutually adjacent surfaces in $\mathcal{A}$ defines the corner
\begin{equation}
\mathbf{v}_{str}=-\begin{bmatrix}\mathbf{n}_s^{\top}\\ \mathbf{n}_t^{\top}\\ \mathbf{n}_r^{\top}\end{bmatrix}^{-1}\begin{bmatrix}d_s\\ d_t\\ d_r\end{bmatrix}.
\label{eq:supp-corner}
\end{equation}
Intersections with the adjacent walls, the floor, and the ceiling define each wall polygon, and neighboring polygons reference the same corner positions.

\paragraph{Junction term.}
The VLM marks visible physical junctions in the source frames and associates each with its two incident surfaces. We sample pixels along each observed junction, excluding foreground occlusions and image-border edges; for junction $k$ in frame $i$, these pixels form $B_{ik}$. For the corresponding plane pair, we compute the 3D intersection line, project two of its points through camera $i$, and take the image line $\boldsymbol{\ell}_{ik}$ through the two projections, normalized so that its first two coefficients have unit Euclidean norm. The inner product $\boldsymbol{\ell}_{ik}^{\top}\tilde{\mathbf{u}}$ is then the signed distance in pixels from $\mathbf{u}$ to the line.

\paragraph{Optimization.}
The active parameters $\boldsymbol{\theta}$ are the plane offsets and corrections to the normal directions, and each may move by at most $\epsilon_m$ from its initial value.
We solve Eq.~\eqref{eq:shell-joint-fit} as a bounded robust least-squares problem with the cameras, observations, and adjacency fixed. At every function evaluation, the normals are renormalized and all shared corners are recomputed with Eq.~\eqref{eq:supp-corner}. The fitted polygons are triangulated into Blender meshes and rendered from additional source views, where we check visible boundaries and point-to-triangle distances. A revision is retained only if it improves the targeted mismatch without opening gaps or degrading boundaries that were already aligned.

\paragraph{Occluded walls.}
For a partially occluded wall, the VLM matches the exposed wall regions across frames and identifies their junctions with neighboring walls, the floor, and the ceiling. We intersect these surfaces with the fitted wall plane and triangulate the resulting polygon across hidden regions, such as the part behind a curtain. The hidden part is marked as inferred, and the visible parts of the extended plane and its boundaries are checked against complementary source views.

\paragraph{Irregular patches.}
For an irregular surface, the VLM identifies visible surface regions and structural boundaries, which determine the support points and connections of each patch. We fit the coefficients $\mathbf{C}$ of Eq.~\eqref{eq:irregular-surface-parameterization} with the reference frames and prescribed coordinate components fixed; planes, periodic lofts, and local quadratic displacement fields (including height fields) are the only representations used, and we do not search a larger primitive library.

\emph{Periodic lofts.} The local coordinates are the horizontal angle and the height, and the horizontal position combines periodic trigonometric functions of the angle with quadratic terms in height. The coefficients are estimated by weighted linear least squares. To constrain poorly observed regions, we add samples from a prescribed elliptical envelope that narrows with height.

\emph{Local quadratic displacement fields.} The reference origin is the coordinate-wise median of the support points. An initially near-vertical patch uses a horizontal reference normal pointing from the mean position of the cameras that observe it toward this origin. With the frame fixed, we fit the normal displacement by linear least squares, penalizing the quadratic coefficients more strongly than the others to limit unsupported curvature. Height fields, obtained with a vertical reference normal, are fitted with the soft-$L_1$ penalty $\rho$ to reduce the influence of depth outliers. When a height field has an observed outer contour, we also adjust its coefficients and the parameters of an elliptical outline so that its projected boundary aligns with traced image curves, bounding and penalizing changes from the initialization.

\paragraph{Patch connections.}
For a sharp junction prescribed by the VLM, we bracket the roots of one surface's implicit equation along the other surface and solve for them with Brent's method. The resulting intersection curve trims both parameter domains. We triangulate the domains and evaluate the surface maps at their vertices; matching boundary samples give corresponding seam vertices identical positions in both patches. Where adjacent patches require a smooth connection, we locally subdivide mesh edges and clip overlapping portions. Each implicit surface function is divided by its gradient norm, and the resulting fields are combined with a smooth maximum. Nearby vertices are projected onto the zero level set of the blended field using finite-difference gradients and bounded steps. The blend width varies spatially and decreases to zero near protected boundaries, which keeps those boundaries fixed.

\subsection{Object Registration}
\label{sec:supp-registration}

\paragraph{Primitive modeling.}
For a large object, the VLM assembles an editable model $M_j^{\mathrm{prim}}$ in Blender through MCP, using simple primitives for the observed major components; their dimensions and relative transforms encode the object's structure and proportions. The VLM places the model roughly: its position and size follow the center and extent of $\mathcal{P}^{\mathrm{obj}}_j$, its base rests on the supporting surface given by $\mathcal{G}$, and its facing direction, which the extent alone does not determine, is read from recognizable components in the frames, such as the side a chair's backrest is on. If the VLM judges that $M_j^{\mathrm{prim}}$ captures the observed components and proportions, it is kept as $M_j$.

\paragraph{Generation with TRELLIS.2.}
Otherwise, we pick three clear, complementary frames from $\mathcal{V}_j$, complete the partially occluded object in each by image editing, and generate one TRELLIS.2 candidate per completed image. Because TRELLIS.2 produces each mesh in its own canonical frame, we first align each candidate's local axes and contact reference and then place it using $M_j^{\mathrm{prim}}$ as the reference for position, orientation, and size. The candidates are rendered from the source cameras, and the VLM keeps the one whose components, proportions, and asymmetric structure best match the original frames; it replaces $M_j^{\mathrm{prim}}$ while keeping the instance's identity and support relations. Small objects skip primitive modeling: TRELLIS.2 generates their meshes directly, and the VLM initializes their position, scale, and orientation from the spatial distribution of $\mathcal{P}^{\mathrm{obj}}_j$ and the source images.
 
\paragraph{Pose parameterization.}
Registration keeps the mesh geometry and the cameras fixed and changes only the transform. For upright furniture, the support fixes the base height, roll, and pitch. Let $\hat{M}_j$ be the mesh divided by its local bounding-box dimensions, with its origin moved to the bottom center. In a frame whose vertical axis is the support normal, the transform is
\begin{equation}
\mathcal{T}_{\mathbf{q}}(\hat{\mathbf{x}})=\mathbf{R}(\psi)\,\mathrm{diag}(\mathbf{e})\,\hat{\mathbf{x}}+\bigl[\mathbf{t}^{\top},\,h_j\bigr]^{\top},
\quad\mathbf{q}=(\mathbf{t},\psi,\mathbf{e}),
\label{eq:object-transform}
\end{equation}
where $\mathbf{t}\in\mathbb{R}^{2}$ is the horizontal translation, $\mathbf{R}(\psi)$ the rotation by yaw $\psi$ about the support normal, $\mathbf{e}\in\mathbb{R}^{3}_{>0}$ the dimensions along the object's local axes, and $h_j$ the base height given by the support. The admissible box $\mathcal{B}_j$ fixes an interval around the initial parameters $\mathbf{q}^{(0)}$ for every parameter: translation limits scale with the initial object extent, yaw has a bounded deviation, and dimension limits are proportional to the initial dimensions. For an upright object on a tabletop, $\mathbf{q}$ contains the two in-plane translations and the rotation about the tabletop normal, and scale is added only if the projected dimensions remain inconsistent after the pose is corrected. All distances are measured in the common reconstruction coordinate system.
 
\paragraph{ICP.}
We sample $N_{\mathrm{surf}}$ points on the surface of $\hat{M}_j$ and build a KD-tree over them. At iteration $t$, each observation $\mathbf{p}_n\in\mathcal{P}^{\mathrm{obj}}_j$ is mapped into normalized model coordinates by $\mathcal{T}^{-1}_{\mathbf{q}^{(t)}}$ and assigned its nearest surface sample $\hat{\mathbf{y}}_n$. With these correspondences fixed, we solve
\begin{equation}
\begin{aligned}
\mathbf{q}^{(t+1)}=\arg\min_{\mathbf{q}\in\mathcal{B}_j}\;
&\sum_{n}\rho_{\delta}\bigl(\mathcal{T}_{\mathbf{q}}(\hat{\mathbf{y}}_n)-\mathbf{p}_n\bigr)\\
&+\rho_{\delta}\bigl(w_{\mathrm{prior}}(\mathbf{q}-\mathbf{q}^{(0)})\bigr),
\end{aligned}
\label{eq:supp-object-pc}
\end{equation}
where $\rho_{\delta}$ applies $\rho(\cdot/\delta)$ to every scalar component of its argument and sums the results, $\delta$ is a fixed soft-$L_1$ scale, and $w_{\mathrm{prior}}$ is a fixed prior weight. Each inner solve has a fixed budget of function evaluations. Correspondence assignment and parameter estimation alternate until the largest absolute parameter change falls below $\epsilon_{\mathrm{icp}}$ or $N_{\mathrm{icp}}$ iterations are reached.
 
\paragraph{VLM-guided adjustment.}
The agent selects three frames from $\mathcal{V}_j$ that together view the object from as many sides as possible and renders the placed object from the same cameras. The VLM compares each rendering with its frame, aided by overlays that mark the area covered only by the rendering and the area covered only by the observation, and adjusts the size and pose within $\mathcal{B}_j$. Small objects occupy few pixels in distant views, so their adjustment uses close-up frames. Each instance is adjusted with its own observations, including when several instances share one generated mesh.
 
\paragraph{Silhouette refinement.}
The adjustment can be made precise with a silhouette objective over all views $i\in\mathcal{V}_j$. Let $A_i$ be the observed instance mask, $H_i(\mathbf{q})$ the visible mask of the object rendered together with the rest of the scene, and $L^{\mathrm{mask}}_i=1-\mathrm{IoU}\bigl(A_i,H_i(\mathbf{q})\bigr)$. We minimize
\begin{equation}
\begin{aligned}
L(\mathbf{q})={}&\frac{1}{|\mathcal{V}_j|}\sum_{i\in\mathcal{V}_j}\bigl(L^{\mathrm{mask}}_i+\lambda_{\mathrm{edge}}E_i\bigr)\\
&+\lambda_{\mathrm{worst}}\max_{i\in\mathcal{V}_j}L^{\mathrm{mask}}_i+\lambda_{\mathrm{pc}}L_{\mathrm{pc}},\\
E_i={}&\frac{\bar{d}(\partial H_i,\partial A_i)+\bar{d}(\partial A_i,\partial H_i)}{2\,\mathrm{diag}(A_i)},
\end{aligned}
\label{eq:object-silhouette}
\end{equation}
where $\lambda_{\mathrm{edge}}$, $\lambda_{\mathrm{worst}}$, and $\lambda_{\mathrm{pc}}$ are fixed weights, $\bar{d}(X,Y)$ is the mean distance from the pixels of $X$ to the nearest pixel of $Y$, $\mathrm{diag}(A_i)$ is the diagonal of the bounding box of $A_i$, and $L_{\mathrm{pc}}$ averages normalized and capped point-to-surface distances. The worst-view term keeps a single badly aligned view from disappearing into the average. To render $H_i(\mathbf{q})$, we ray-cast the target object and the rest of the scene from the recovered camera and keep the target pixels that are not occluded by other scene geometry, up to a numerical depth tolerance $\tau_{\mathrm{depth}}$. Masks are evaluated at a reduced resolution with correspondingly scaled intrinsics. The boundary $\partial X$ of a mask $X$ is obtained by subtracting its eroded version, and image-border pixels are excluded. Euclidean distance transforms give the mean nearest-boundary distances $\bar{d}$ in both directions; the normalizer $\mathrm{diag}(A_i)$ is computed once at the reduced resolution, and an empty rendered boundary receives a fixed penalty $E_{\mathrm{empty}}$. For $L_{\mathrm{pc}}$, each world-space correspondence distance is divided by a distance scale $d_{\mathrm{pc}}$, capped at $\tau_{\mathrm{cap}}$, and averaged over the observations; correspondences are recomputed with the same normalized-space nearest-neighbor rule as in ICP, using a separate surface sample of the mesh.
 
\paragraph{Coordinate search.}
Because $L$ is built from binary masks and is not differentiable in $\mathbf{q}$, we minimize it by bounded coordinate search. The search starts from whichever of $\mathbf{q}^{(0)}$ and the ICP solution has the lower $L$. The translation, yaw, and lower dimension bounds of $\mathcal{B}_j$ are kept, while the upper dimension bounds are relaxed by a predefined amount $\beta_{\mathrm{dim}}$. Translation and dimension steps are proportional to the initial object dimensions, and yaw uses a fixed angular step. For each coordinate in turn, we evaluate a positive and a negative step, both clipped to the bounds, by re-rendering all visible masks and recomputing $L$, and we keep the better trial only if it reduces $L$ by more than $\epsilon_{\mathrm{cs}}$. Step sizes decrease over $N_{\mathrm{lvl}}$ search levels, each with a budget of $N_{\mathrm{sweep}}$ sweeps, and a sweep without improvement advances the search to the next level.
 
\paragraph{Final check.}
The accepted transform is applied in Blender and re-rendered from the source cameras. We recompute point-to-surface distances against the full triangle mesh and inspect the overlays before retaining the placement.

\subsection{Part-Wise Refinement}
\label{sec:supp-part}

The VLM compares the source views with the current renderings to identify a component that the object-level transform cannot fit, for example cushions that remain misplaced once the sofa body matches, or an overly wide backrest whose seat and armrests already match. To select the component, the agent marks its visible regions in multiple renderings of the current object, and camera rays through these regions are intersected with the mesh to find the corresponding triangles. The selection is highlighted in complementary model views so that the agent can check that it covers the intended component and excludes neighboring parts, and additional views are used to add side and rear surfaces and to correct selection errors. The vertices of the reviewed selection form a vertex group $\mathcal{U}$; the mesh is neither cut nor split into a separate object. These rendered annotations only determine which existing geometry is editable; the source frames remain the reference for the target shape.
 
For a stretch, the fixed anchor $\mathbf{x}_{\mathrm{fix}}$ is placed where the component's alignment or attachment should be preserved, and the moving anchor $\mathbf{x}_{\mathrm{mov}}$ toward the extent that needs correction. Corresponding landmarks in the current renderings are ray-cast onto the mesh to locate both anchors in object-local coordinates. The agent estimates $\Delta$ from the source boundaries and the available reliable depth, and checks the proposed change in complementary source views before applying Eq.~\eqref{eq:part-stretch} to the vertices in $\mathcal{U}$. After each edit, we re-render from the relevant source cameras and assess the component's boundary alignment, its connection to neighboring components, and its support contact. The edit is kept only if it improves the targeted discrepancy without degrading other components or introducing severe interpenetration; otherwise, the previous geometry is restored.

\subsection{Cross-Object Refinement}
\label{sec:supp-cross}
 
For each support relation $(u,o_j)\in\mathcal{E}$, the object is translated along the support normal until its base touches the mesh of $u$, which stays fixed. Several samples across the intended contact region are then checked so that the object neither floats above nor penetrates its support. Wall-mounted and suspended objects are constrained by their observed attachment regions instead. Each correction is also checked in complementary views and against neighboring objects for spacing and occlusion. Supports are processed before the objects they support; objects on different supports are processed independently, and a later change to a support requires re-checking the objects it supports.

\subsection{Agent}
\label{sec:supp-agent}

The skill specification defines the stage dependencies, the tools available at each stage, and the criteria a candidate must meet to be accepted. Each review records the localized discrepancies and the corrective actions taken, and every accepted scene state is stored so that later revisions can be rolled back to it.
Both reconstruction stages are executed by a tool-using VLM agent that follows a reusable skill specification defining the stage dependencies, available operations, and verification criteria (Appendix~\ref{sec:supp-agent}). Rather than producing geometry itself, the agent works through tool calls: frame extraction, DA3 inference, image editing, TRELLIS.2 generation, and Blender operations through MCP. Numerical tools perform the shell fitting, object registration, mesh measurements, and mask comparisons, while the agent makes the decisions they cannot: which representation a surface needs, which frames to use, which component is wrong, and whether a result matches the video. Following the design--critique--orchestration organization of SceneSmith~\cite{pfaff2026scenesmith}, we describe this work in three roles.
 
\paragraph{Construction.}
The agent inspects the video, selects informative views, and invokes the modeling and generation tools: it initializes the shell and chooses the parameters to optimize, prepares generation inputs, compares candidate assets in the scene, and adjusts their placement.
 
\paragraph{Verification.}
The agent compares source frames with textured and textureless renderings from the same cameras, together with overlays of extra and missing regions and measurements of mask overlap, boundary distance, and point-to-surface distance. It checks structural completeness, object identity, proportions, orientation, and support and occlusion relations. Rigid major structures must agree at their boundaries, whereas soft or intricate components are judged by overall volume, color, and arrangement. Only the original frames serve as the reference; edited completions are used solely to generate candidates. Each discrepancy is localized to one component and mapped to an operation: constrained transforms for pose and scale errors, part-wise deformation for local shape errors, surface refitting for shell mismatches, geometry correction or candidate replacement for structural errors, and material adjustment for appearance errors.
 
\paragraph{Orchestration.}
Each revision yields a candidate scene that is evaluated from the relevant source views and accepted only if it improves the diagnosed discrepancy without regressions in other views; a lower aggregate error cannot override a missing structure, severe interpenetration, or incorrect support. Otherwise, the last accepted state is restored. The shell is reviewed before any furniture is built, and a change to a supporting object triggers new checks of everything it supports. When the evidence is ambiguous, the agent searches the video for clearer or complementary views before editing further.

\begin{table}[t]
\centering
\small
\caption{\textbf{Constants used by \nickname{}.} Symbols follow Sec.~\ref{sec:method} and Sec.~\ref{sec:supp-impl}.}
\label{tab:supp-hparams}
\resizebox{\linewidth}{!}{%
\begin{tabular}{p{0.56\linewidth}cc}
\hline
Quantity & Symbol & Value \\
\hline
\multicolumn{3}{l}{\emph{Scene preprocessing}}\\
Frames sampled per video (real / simulated) & $T$ & 100 / 60 \\
Mask erosion kernel & $r_{\mathrm{erode}}$ & $3\times3$ \\
Confidence floor / quantile & $\tau_{\mathrm{conf}}$, $\kappa$ & $1.2 / 0.35$ \\
Points per view & $N_{\mathrm{view}}$ & $650$ \\
\hline
\multicolumn{3}{l}{\emph{Object generation}}\\
Source views per large object & -- & 3 \\
TRELLIS.2 candidates per view & -- & 1 \\
\hline
\multicolumn{3}{l}{\emph{Object registration}}\\
Translation / yaw / dimension bounds of $\mathcal{B}_j$ & -- & \shortstack{$\pm0.18\max_k e_k^{(0)}$\\$\pm0.45\,\mathrm{rad}$\\$[0.75,1.25]\mathbf{e}^{(0)}$} \\
Surface samples & $N_{\mathrm{surf}}$ & \shortstack{$70{,}000$ (ICP)\\$60{,}000$ (silhouette)} \\
Soft-$L_1$ scale / prior weight & $\delta$, $w_{\mathrm{prior}}$ & $0.025 / 0.03$ \\
ICP tolerance / max.\ iterations & $\epsilon_{\mathrm{icp}}$, $N_{\mathrm{icp}}$ & $10^{-5} / 25$ \\
Function evaluations per inner solve & -- & $50$ \\
Mask resolution factor & -- & $0.25$ \\
Depth tolerance & $\tau_{\mathrm{depth}}$ & $10^{-5}$ \\
Empty-boundary penalty & $E_{\mathrm{empty}}$ & $1$ \\
Silhouette weights & $\lambda_{\mathrm{edge}}$, $\lambda_{\mathrm{worst}}$, $\lambda_{\mathrm{pc}}$ & $0.4 / 0.35 / 0.12$ \\
Point-distance scale / cap & $d_{\mathrm{pc}}$, $\tau_{\mathrm{cap}}$ & $0.05 / 2$ \\
Upper-dimension relaxation & $\beta_{\mathrm{dim}}$ & $0.05$ \\
Initial translation / dimension / yaw steps & -- & \shortstack{$0.035\max_k e_k^{(0)}$\\$0.045\mathbf{e}^{(0)}$\\$0.045\,\mathrm{rad}$} \\
Search levels / sweeps per level & $N_{\mathrm{lvl}}$, $N_{\mathrm{sweep}}$ & $3 / 4$ \\
Step schedule across levels & -- & $1,0.5,0.25$ \\
Acceptance tolerance & $\epsilon_{\mathrm{cs}}$ & $10^{-5}$ \\
\hline
\multicolumn{3}{l}{\emph{Agent}}\\
Max.\ revision rounds per stage & -- & $3$ \\
\hline
\end{tabular}
}
\end{table}

\subsection{Evaluation Protocol}
\label{sec:supp-evaluation}

\paragraph{Shared scene sets.}
The baseline comparisons in Sec.~\ref{sec:experiments} use a fixed set of eight Replica scenes and a fixed set of fifteen Imaginarium scenes. For each dataset, the evaluation set is the union of the scenes used across the baseline runs, and GRASE, SAM3D, GPT-6 vanilla, and GPT-6 + TRELLIS.2 + point cloud are all evaluated on this entire set. Every method successfully completed reconstruction on every scene. The reported baseline comparisons therefore do not use different scene subsets for different methods.

\paragraph{Object-level denominator.}
For a given scene, all methods are evaluated against the same full set of GT objects. The denominator for each reported object-level metric (mIoU, NC, CD, and F@5cm) is the full GT object count, not the number of predicted, successfully reconstructed, or matched objects. GT objects are not removed from this denominator based on a method's reconstruction or matching outcome.

\paragraph{Rendering availability.}
The missing SAM3D rendering metrics in Table~\ref{tab:main_results} reflect the absence of authored lighting and a world setup in its delivered GLB files, not failed reconstructions or a smaller evaluation scene set.

\paragraph{Shell ablation protocol.}
The point-cloud and hard-crease critic ablation in Table~\ref{tab:shell_ablation}b is averaged equally over seven scenes after upright rigid alignment with fixed scale. CD is reported in $10^{-2}$ source-coordinate units, and F-score uses a threshold of $0.05$ source-coordinate units.

%% file: comparison_multiview/four_scene_multiview_full.tex
\begin{figure*}[p]
\centering
\includegraphics[width=\textwidth,height=0.87\textheight,keepaspectratio]{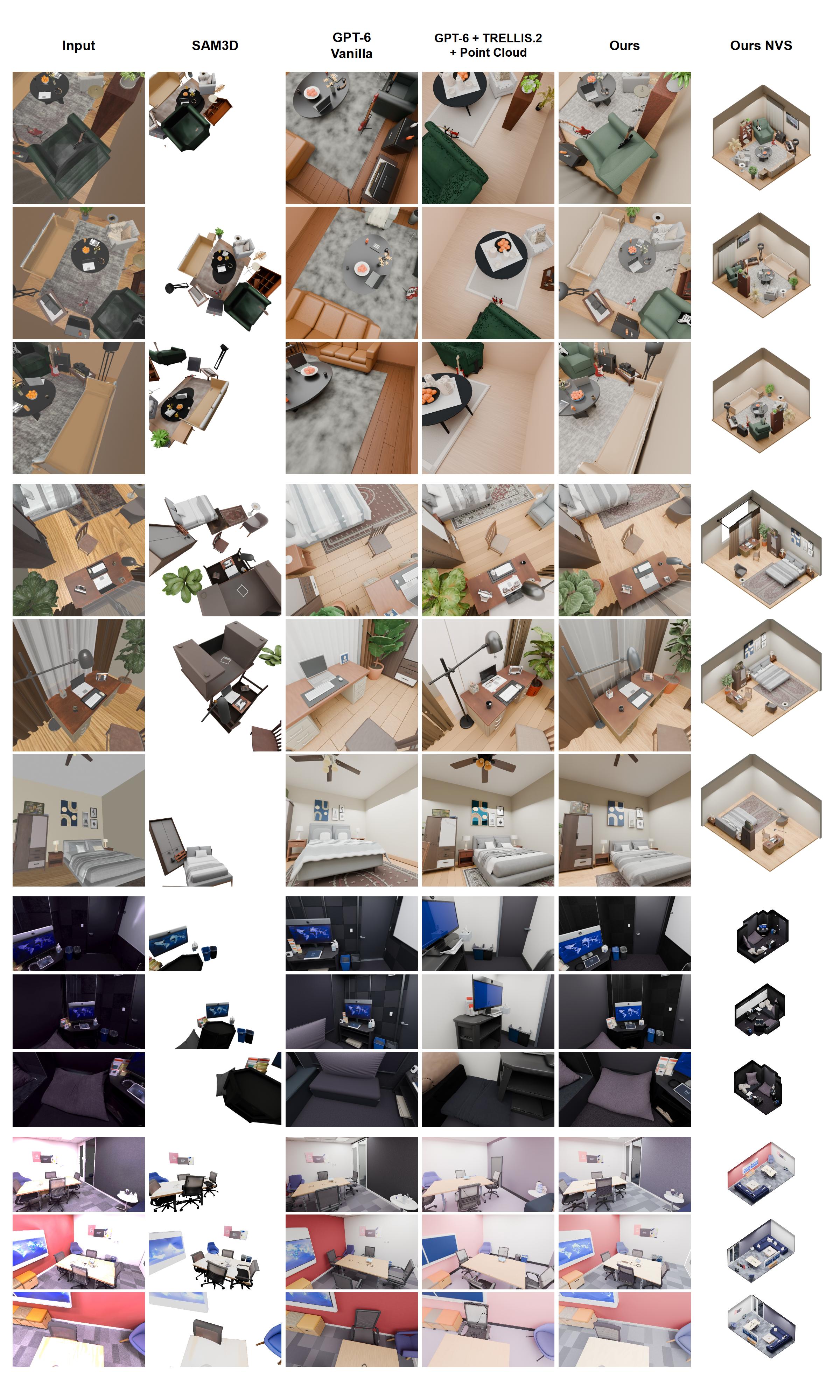}
\caption{Full qualitative multiview comparison for the four scenes of Figure~\ref{fig:qualitative_four_scene_comparison}. Each group of three rows shows one scene from three input-camera viewpoints. The first five columns share camera poses and intrinsics within each row; Ours NVS shows Blender scene overviews.}
\label{fig:qualitative_full_multiview}
\end{figure*}

%% file: appendix_qualitative/tex/gallery.tex
\clearpage
\section{Additional Qualitative Results}
\label{sec:appendix-qualitative}

Figures~\ref{fig:appendix-ours-portrait}--\ref{fig:appendix-imaginarium} show reconstructions of 23 scenes: seven from our scene collection, eight from Replica, and eight from Imaginarium.
These figures illustrate selected reconstructions rather than define the quantitative evaluation sets. In particular, the eight Imaginarium scenes shown here are a subset of the fifteen scenes used by every method in the baseline comparisons; all methods use the same eight Replica and fifteen Imaginarium evaluation scenes (Appendix~\ref{sec:supp-evaluation}).

\begingroup
\newcommand{\appSceneHeader}{%
  \noindent\makebox[.162\linewidth]{\scriptsize Input}\hfill
  \makebox[.162\linewidth]{\scriptsize Render 1}\hfill
  \makebox[.162\linewidth]{\scriptsize Render 2}\hfill
  \makebox[.162\linewidth]{\scriptsize Render 3}\hfill
  \makebox[.162\linewidth]{\scriptsize Render 4}\hfill
  \makebox[.162\linewidth]{\scriptsize Render 5}\par\vspace{3pt}%
}
\newsavebox{\appCompositeBox}

\subsection{Our Scene Collection}
\begin{figure}[!ht]
  \centering
  \sbox{\appCompositeBox}{%
    \includegraphics[width=\linewidth,height=0.66\textheight,keepaspectratio]{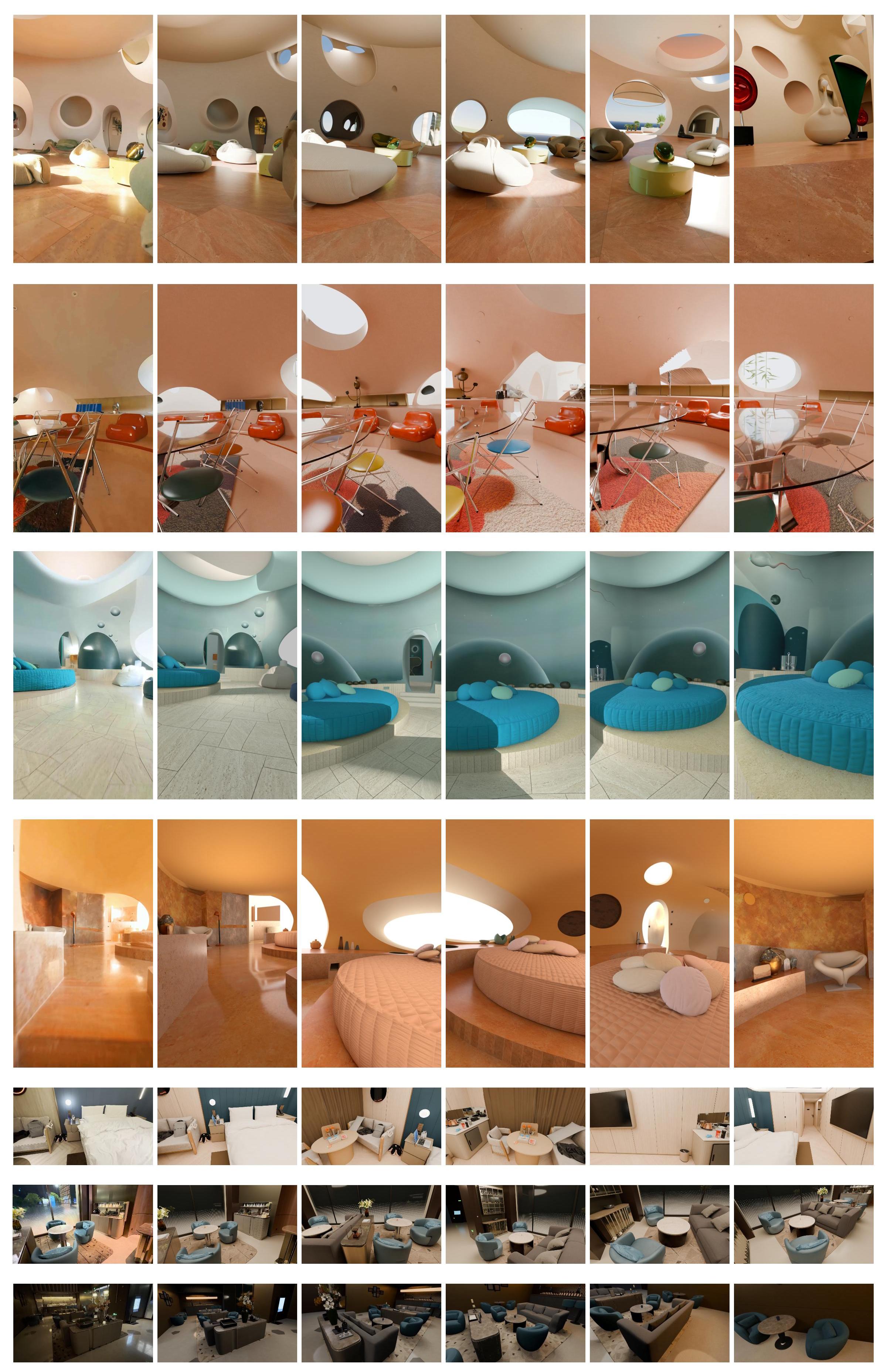}}%
  \begin{minipage}{\wd\appCompositeBox}
    \appSceneHeader
    \usebox{\appCompositeBox}
  \end{minipage}
  \caption{\textbf{Our scene collection.} Each row shows an input frame followed by five renderings of the reconstructed scene.}
  \label{fig:appendix-ours-portrait}
  \label{fig:appendix-ours-landscape}
\end{figure}

\clearpage
\subsection{Replica Scenes}
\begin{figure}[!ht]
  \centering
  \sbox{\appCompositeBox}{%
    \includegraphics[width=\linewidth,height=0.80\textheight,keepaspectratio]{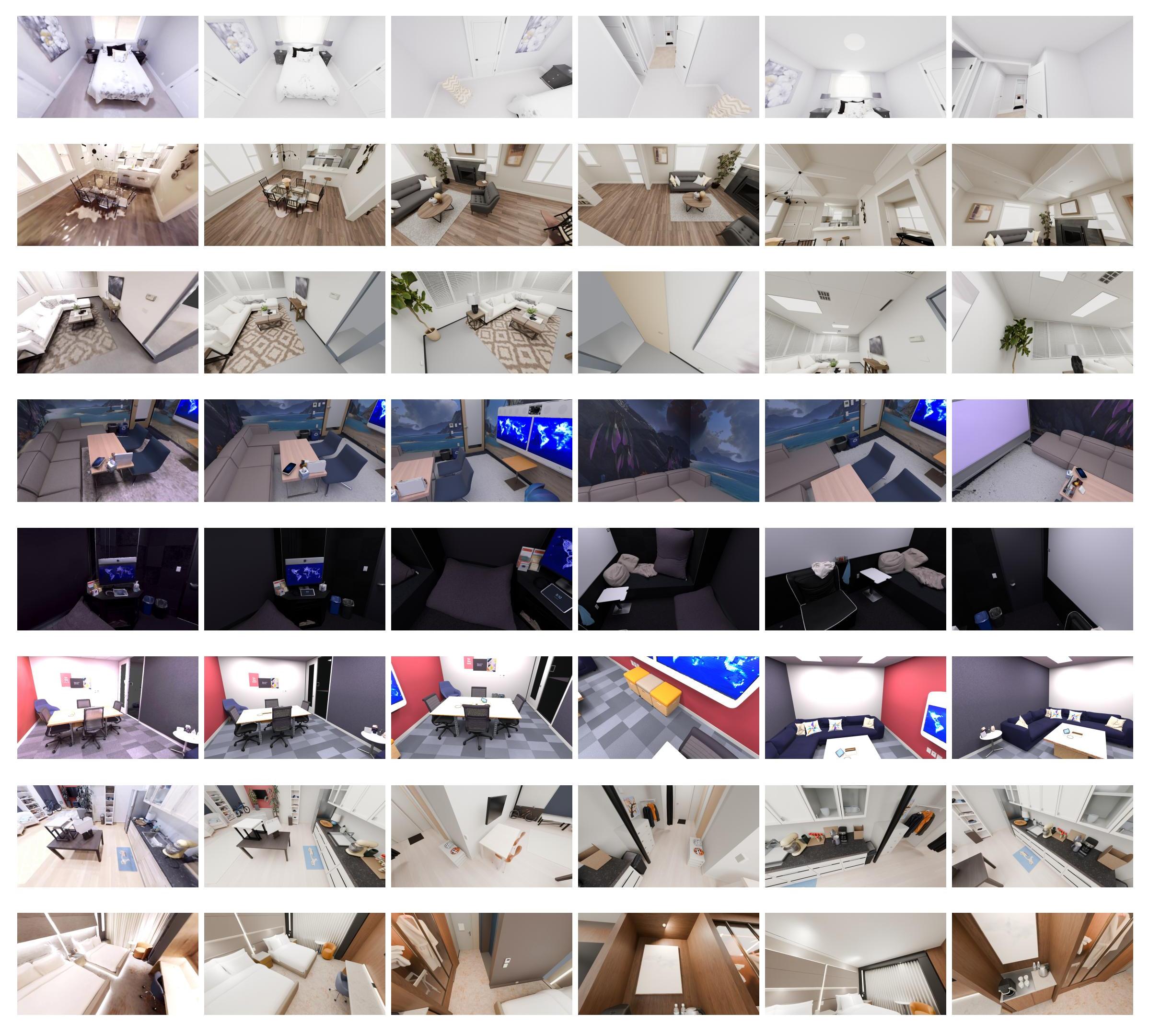}}%
  \begin{minipage}{\wd\appCompositeBox}
    \appSceneHeader
    \usebox{\appCompositeBox}
  \end{minipage}
  \caption{\textbf{Replica scenes.} Columns follow Figure~\ref{fig:appendix-ours-portrait}.}
  \label{fig:appendix-replica}
\end{figure}

\clearpage
\subsection{Imaginarium Scenes}
\label{sec:appendix-imaginarium}
\begin{figure}[!ht]
  \centering
  \sbox{\appCompositeBox}{%
    \includegraphics[width=\linewidth,height=0.80\textheight,keepaspectratio]{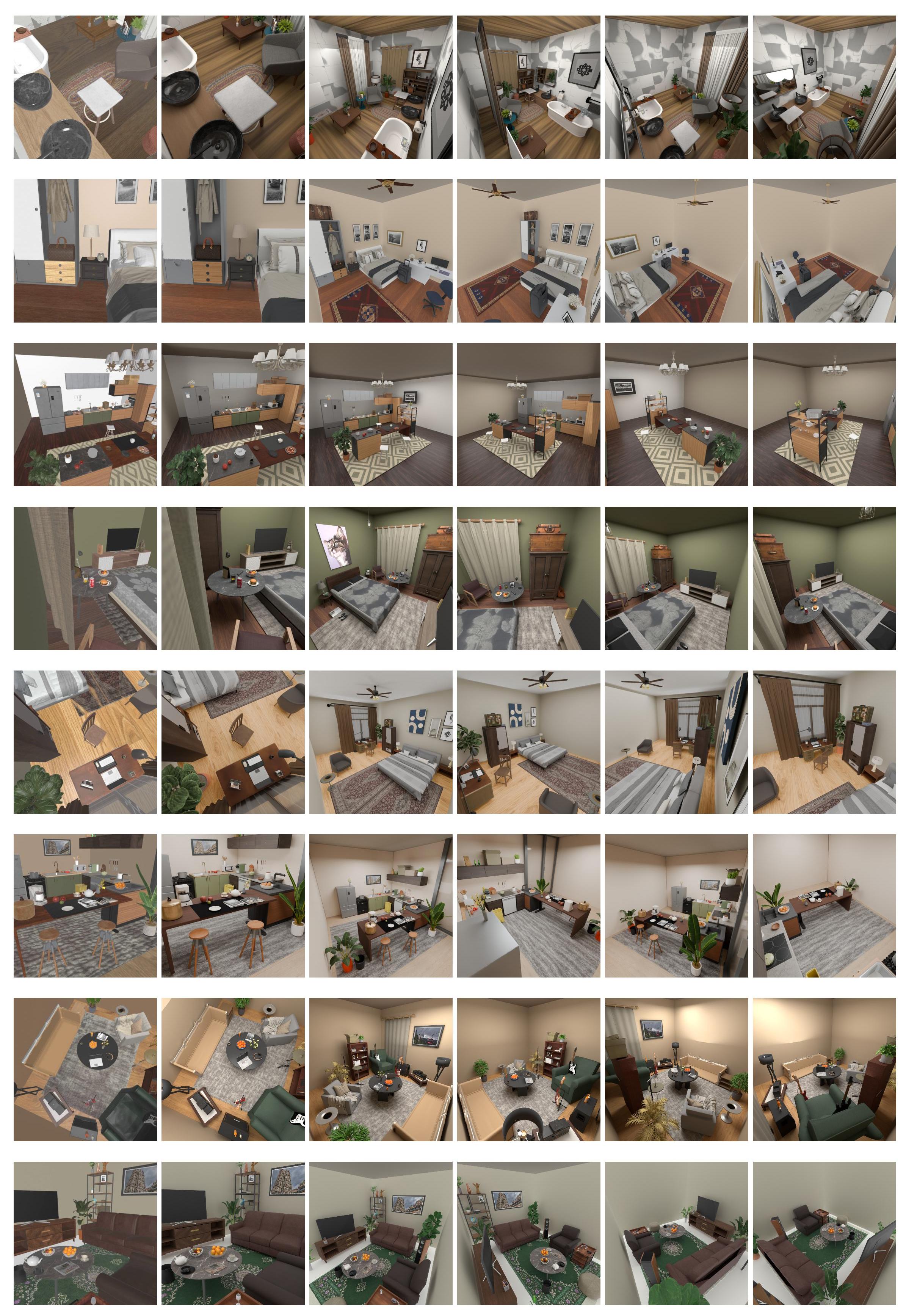}}%
  \begin{minipage}{\wd\appCompositeBox}
    \appSceneHeader
    \usebox{\appCompositeBox}
  \end{minipage}
  \caption{\textbf{Imaginarium scenes.} Columns follow Figure~\ref{fig:appendix-ours-portrait}.}
  \label{fig:appendix-imaginarium}
  \label{fig:appendix-imaginarium-remaining}
\end{figure}
\clearpage
\endgroup

%% file: main.bbl
\begin{thebibliography}{37}
\providecommand{\natexlab}[1]{#1}
\providecommand{\url}[1]{\texttt{#1}}
\expandafter\ifx\csname urlstyle\endcsname\relax
  \providecommand{\doi}[1]{doi: #1}\else
  \providecommand{\doi}{doi: \begingroup \urlstyle{rm}\Url}\fi

\bibitem[Cai et~al.(2025)Cai, Ponomarenko, Yuan, Li, Yang, Dong, and Zhao]{cai2025spatialbot}
Wenxiao Cai, Iaroslav Ponomarenko, Jianhao Yuan, Xiaoqi Li, Wankou Yang, Hao Dong, and Bo~Zhao.
\newblock Spatialbot: Precise spatial understanding with vision language models.
\newblock In \emph{2025 IEEE International Conference on Robotics and Automation (ICRA)}, pp.\  9490--9498. IEEE, 2025.

\bibitem[{\c{C}}elen et~al.(2024){\c{C}}elen, Han, Schindler, Van~Gool, Armeni, Obukhov, and Wang]{ccelen2024design}
Ata {\c{C}}elen, Guo Han, Konrad Schindler, Luc Van~Gool, Iro Armeni, Anton Obukhov, and Xi~Wang.
\newblock I-design: Personalized llm interior designer.
\newblock In \emph{European Conference on Computer Vision}, pp.\  217--234. Springer, 2024.

\bibitem[Chen et~al.(2024{\natexlab{a}})Chen, Xu, Kirmani, Ichter, Sadigh, Guibas, and Xia]{chen2024spatialvlm}
Boyuan Chen, Zhuo Xu, Sean Kirmani, Brain Ichter, Dorsa Sadigh, Leonidas Guibas, and Fei Xia.
\newblock Spatialvlm: Endowing vision-language models with spatial reasoning capabilities.
\newblock In \emph{Proceedings of the IEEE/CVF conference on computer vision and pattern recognition}, pp.\  14455--14465, 2024{\natexlab{a}}.

\bibitem[Chen et~al.(2026)Chen, Zhang, Chen, Zhang, Zhang, Zhu, Zhao, Huang, Li, and Wang]{chen2026one}
Junhao Chen, Boran Zhang, Mingjin Chen, Henghaofan Zhang, Saining Zhang, Congcong Zhu, Hao Zhao, Ruqi Huang, Zhihao Li, and Yufei Wang.
\newblock One video, one world: Turning monocular video into physical 4d scenes.
\newblock \emph{arXiv preprint arXiv:2606.31388}, 2026.

\bibitem[Chen et~al.(2024{\natexlab{b}})Chen, Chen, Zhang, Li, Yu, Fei, Zhu, Fan, and Chen]{chen2024ll3da}
Sijin Chen, Xin Chen, Chi Zhang, Mingsheng Li, Gang Yu, Hao Fei, Hongyuan Zhu, Jiayuan Fan, and Tao Chen.
\newblock Ll3da: Visual interactive instruction tuning for omni-3d understanding, reasoning, and planning.
\newblock In \emph{2024 IEEE/CVF Conference on Computer Vision and Pattern Recognition (CVPR)}, pp.\  26418--26428. IEEE, 2024{\natexlab{b}}.

\bibitem[Cheng et~al.(2024)Cheng, Yin, Fu, Guo, Yang, Kautz, Wang, and Liu]{cheng2024spatialrgpt}
An-Chieh Cheng, Hongxu Yin, Yang Fu, Qiushan Guo, Ruihan Yang, Jan Kautz, Xiaolong Wang, and Sifei Liu.
\newblock Spatialrgpt: Grounded spatial reasoning in vision-language models.
\newblock \emph{Advances in Neural Information Processing Systems}, 37:\penalty0 135062--135093, 2024.

\bibitem[Cui et~al.(2026)Cui, Wang, Zhang, He, Lin, Zhao, He, Chen, Chen, and Zhuang]{cui2026block3d}
Bowen Cui, Weijie Wang, Zeyu Zhang, Yefei He, Mingda Lin, Haoyu Zhao, Yuanyu He, Donny~Y. Chen, Feng Chen, and Bohan Zhuang.
\newblock Block3d: Efficient text-to-3d generation via block-wise diffusion.
\newblock \emph{arXiv preprint arXiv:2608.19567}, 2026.

\bibitem[Dahnert et~al.(2021)Dahnert, Hou, Nie{\ss}ner, and Dai]{dahnert2021panoptic}
Manuel Dahnert, Ji~Hou, Matthias Nie{\ss}ner, and Angela Dai.
\newblock Panoptic 3d scene reconstruction from a single rgb image.
\newblock \emph{Advances in Neural Information Processing Systems}, 34:\penalty0 8282--8293, 2021.

\bibitem[Dahnert et~al.(2024)Dahnert, Dai, M{\"u}ller, and Nie{\ss}ner]{dahnert2024coherent}
Manuel Dahnert, Angela Dai, Norman M{\"u}ller, and Matthias Nie{\ss}ner.
\newblock Coherent 3d scene diffusion from a single rgb image.
\newblock \emph{Advances in Neural Information Processing Systems}, 37:\penalty0 23435--23463, 2024.

\bibitem[Feng et~al.(2023)Feng, Zhu, Fu, Jampani, Akula, He, Basu, Wang, and Wang]{feng2023layoutgpt}
Weixi Feng, Wanrong Zhu, Tsu-jui Fu, Varun Jampani, Arjun Akula, Xuehai He, Sugato Basu, Xin~Eric Wang, and William~Yang Wang.
\newblock Layoutgpt: Compositional visual planning and generation with large language models.
\newblock \emph{Advances in neural information processing systems}, 36:\penalty0 18225--18250, 2023.

\bibitem[Fu et~al.(2024)Fu, Hu, Li, Feng, Wang, Lin, Roth, Smith, Ma, and Krishna]{fu2024blink}
Xingyu Fu, Yushi Hu, Bangzheng Li, Yu~Feng, Haoyu Wang, Xudong Lin, Dan Roth, Noah~A Smith, Wei-Chiu Ma, and Ranjay Krishna.
\newblock Blink: Multimodal large language models can see but not perceive.
\newblock In \emph{European Conference on Computer Vision}, pp.\  148--166. Springer, 2024.

\bibitem[Hong et~al.(2023)Hong, Zhen, Chen, Zheng, Du, Chen, and Gan]{hong20233d}
Yining Hong, Haoyu Zhen, Peihao Chen, Shuhong Zheng, Yilun Du, Zhenfang Chen, and Chuang Gan.
\newblock 3d-llm: Injecting the 3d world into large language models.
\newblock \emph{Advances in Neural Information Processing Systems}, 36:\penalty0 20482--20494, 2023.

\bibitem[Hu et~al.(2024)Hu, Iscen, Jain, Kipf, Yue, Ross, Schmid, and Fathi]{hu2024scenecraft}
Ziniu Hu, Ahmet Iscen, Aashi Jain, Thomas Kipf, Yisong Yue, David~A Ross, Cordelia Schmid, and Alireza Fathi.
\newblock Scenecraft: An llm agent for synthesizing 3d scenes as blender code.
\newblock In \emph{ICML}, pp.\  19252--19282, 2024.

\bibitem[Huang et~al.(2023)Huang, Chen, Wang, Huang, Xu, Wang, Liu, Cheng, Zhao, Pang, et~al.]{huang2023chat}
Haifeng Huang, Yilun Chen, Zehan Wang, Rongjie Huang, Runsen Xu, Tai Wang, Luping Liu, Xize Cheng, Yang Zhao, Jiangmiao Pang, et~al.
\newblock Chat-scene: Bridging 3d scene and large language models with object identifiers.
\newblock \emph{arXiv preprint arXiv:2312.08168}, 2023.

\bibitem[Huang et~al.(2025)Huang, Guo, An, Yang, Li, Zou, Liang, Liu, Cao, and Sheng]{huang2025midi}
Zehuan Huang, Yuan-Chen Guo, Xingqiao An, Yunhan Yang, Yangguang Li, Zi-Xin Zou, Ding Liang, Xihui Liu, Yan-Pei Cao, and Lu~Sheng.
\newblock Midi: Multi-instance diffusion for single image to 3d scene generation.
\newblock In \emph{2025 IEEE/CVF Conference on Computer Vision and Pattern Recognition (CVPR)}, pp.\  23646--23657. IEEE, 2025.

\bibitem[Leroy et~al.(2024)Leroy, Cabon, and Revaud]{leroy2024grounding}
Vincent Leroy, Yohann Cabon, and J{\'e}r{\^o}me Revaud.
\newblock Grounding image matching in 3d with mast3r.
\newblock In \emph{European conference on computer vision}, pp.\  71--91. Springer, 2024.

\bibitem[Ling et~al.(2026)Ling, Lin, Lin, Ding, Zeng, Sheng, Ge, Liu, Bera, and Li]{ling2026scenethesis}
Lu~Ling, Chen-Hsuan Lin, Tsung-Yi Lin, Yifan Ding, Yu~Zeng, Yichen Sheng, Yunhao Ge, Ming-Yu Liu, Aniket Bera, and Max Li.
\newblock Scenethesis: A language and vision agentic framework for 3d scene generation.
\newblock In \emph{International Conference on Learning Representations}, volume 2026, pp.\  136596--136629, 2026.

\bibitem[Liu et~al.(2025)Liu, Tai, and Tang]{liu2025agentic}
Xinhang Liu, Yu-Wing Tai, and Chi-Keung Tang.
\newblock Agentic 3d scene generation with spatially contextualized vlms.
\newblock \emph{arXiv preprint arXiv:2505.20129}, 2025.

\bibitem[Ma et~al.(2024)Ma, Lu, Cheng, Trigoni, and Markham]{ma2024spatialpin}
Chenyang Ma, Kai Lu, Ta-Ying Cheng, Niki Trigoni, and Andrew Markham.
\newblock Spatialpin: Enhancing spatial reasoning capabilities of vision-language models through prompting and interacting 3d priors.
\newblock \emph{Advances in neural information processing systems}, 37:\penalty0 68803--68832, 2024.

\bibitem[Nie et~al.(2020)Nie, Han, Guo, Zheng, Chang, and Zhang]{nie2020total3dunderstanding}
Yinyu Nie, Xiaoguang Han, Shihui Guo, Yujian Zheng, Jian Chang, and Jian~Jun Zhang.
\newblock Total3dunderstanding: Joint layout, object pose and mesh reconstruction for indoor scenes from a single image.
\newblock In \emph{2020 IEEE/CVF Conference on Computer Vision and Pattern Recognition (CVPR)}, pp.\  52--61. IEEE, 2020.

\bibitem[Pfaff et~al.(2026)Pfaff, Cohn, Zakharov, Cory, and Tedrake]{pfaff2026scenesmith}
Nicholas Pfaff, Thomas Cohn, Sergey Zakharov, Rick Cory, and Russ Tedrake.
\newblock Scenesmith: Agentic generation of simulation-ready indoor scenes.
\newblock In \emph{Forty-third International Conference on Machine Learning}, 2026.

\bibitem[Shan et~al.(2026)Shan, Luo, Li, Zheng, Fu, Chen, and Huang]{shan2026interact3d}
Hui Shan, Keyang Luo, Ming Li, Sizhe Zheng, Yanwei Fu, Zhen Chen, and Xiangru Huang.
\newblock Interact3d: Compositional 3d generation of interactive objects.
\newblock \emph{arXiv preprint arXiv:2603.16085}, 2026.

\bibitem[Sun et~al.(2025)Sun, Liu, Gu, Lim, Bhat, Tombari, Li, Haber, and Wu]{sun2025layoutvlm}
Fan-Yun Sun, Weiyu Liu, Siyi Gu, Dylan Lim, Goutam Bhat, Federico Tombari, Manling Li, Nick Haber, and Jiajun Wu.
\newblock Layoutvlm: Differentiable optimization of 3d layout via vision-language models.
\newblock In \emph{2025 IEEE/CVF Conference on Computer Vision and Pattern Recognition (CVPR)}, pp.\  29469--29478. IEEE, 2025.

\bibitem[Wang \& Agapito(2025)Wang and Agapito]{wang20253d}
Hengyi Wang and Lourdes Agapito.
\newblock 3d reconstruction with spatial memory.
\newblock In \emph{2025 International Conference on 3D Vision (3DV)}, pp.\  78--89. IEEE, 2025.

\bibitem[Wang et~al.(2025)Wang, Chen, Karaev, Vedaldi, Rupprecht, and Novotny]{wang2025vggt}
Jianyuan Wang, Minghao Chen, Nikita Karaev, Andrea Vedaldi, Christian Rupprecht, and David Novotny.
\newblock Vggt: Visual geometry grounded transformer.
\newblock In \emph{2025 IEEE/CVF Conference on Computer Vision and Pattern Recognition (CVPR)}, pp.\  5294--5306. IEEE, 2025.

\bibitem[Wang et~al.(2024{\natexlab{a}})Wang, Leroy, Cabon, Chidlovskii, and Revaud]{wang2024dust3r}
Shuzhe Wang, Vincent Leroy, Yohann Cabon, Boris Chidlovskii, and Jerome Revaud.
\newblock Dust3r: Geometric 3d vision made easy.
\newblock In \emph{2024 IEEE/CVF Conference on Computer Vision and Pattern Recognition (CVPR)}, pp.\  20697--20709. IEEE, 2024{\natexlab{a}}.

\bibitem[Wang et~al.(2024{\natexlab{b}})Wang, Qiu, Liu, Chen, Cai, Wang, Wang, Xian, and Gan]{wang2024architect}
Yian Wang, Xiaowen Qiu, Jiageng Liu, Zhehuan Chen, Jiting Cai, Yufei Wang, Tsun-Hsuan Wang, Zhou Xian, and Chuang Gan.
\newblock Architect: Generating vivid and interactive 3d scenes with hierarchical 2d inpainting.
\newblock \emph{Advances in Neural Information Processing Systems}, 37:\penalty0 67575--67603, 2024{\natexlab{b}}.

\bibitem[Wen et~al.(2023)Wen, Liu, Sridhar, and Fu]{wen2023anyhome}
Zehao Wen, Zichen Liu, Srinath Sridhar, and Rao Fu.
\newblock Anyhome: Open-vocabulary generation of structured and textured 3d homes.
\newblock \emph{arXiv e-prints}, pp.\  arXiv--2312, 2023.

\bibitem[Xia et~al.(2026)Xia, Li, Li, Ma, Xu, Liu, Cui, Lin, Ma, Wang, et~al.]{xia2026sage}
Hongchi Xia, Xuan Li, Zhaoshuo Li, Qianli Ma, Jiashu Xu, Ming-Yu Liu, Yin Cui, Tsung-Yi Lin, Wei-Chiu Ma, Shenlong Wang, et~al.
\newblock Sage: Scalable agentic 3d scene generation for embodied ai.
\newblock \emph{arXiv preprint arXiv:2602.10116}, 2026.

\bibitem[Yang et~al.(2025{\natexlab{a}})Yang, Sax, Liang, Henaff, Tang, Cao, Chai, Meier, and Feiszli]{yang2025fast3r}
Jianing Yang, Alexander Sax, Kevin~J Liang, Mikael Henaff, Hao Tang, Ang Cao, Joyce Chai, Franziska Meier, and Matt Feiszli.
\newblock Fast3r: Towards 3d reconstruction of 1000+ images in one forward pass.
\newblock In \emph{2025 IEEE/CVF Conference on Computer Vision and Pattern Recognition (CVPR)}, pp.\  21924--21935. IEEE, 2025{\natexlab{a}}.

\bibitem[Yang et~al.(2025{\natexlab{b}})Yang, Yang, Gupta, Han, Fei-Fei, and Xie]{yang2025thinking}
Jihan Yang, Shusheng Yang, Anjali~W Gupta, Rilyn Han, Li~Fei-Fei, and Saining Xie.
\newblock Thinking in space: How multimodal large language models see, remember, and recall spaces.
\newblock In \emph{2025 IEEE/CVF Conference on Computer Vision and Pattern Recognition (CVPR)}, pp.\  10632--10643. IEEE, 2025{\natexlab{b}}.

\bibitem[Yang et~al.(2026)Yang, Jia, Zhang, and Huang]{yang2026sceneweaver}
Yandan Yang, Baoxiong Jia, Shujie Zhang, and Siyuan Huang.
\newblock Sceneweaver: All-in-one 3d scene synthesis with an extensible and self-reflective agent.
\newblock \emph{Advances in neural information processing systems}, 38:\penalty0 140319--140351, 2026.

\bibitem[Yang et~al.(2024)Yang, Sun, Weihs, VanderBilt, Herrasti, Han, Wu, Haber, Krishna, Liu, et~al.]{yang2024holodeck}
Yue Yang, Fan-Yun Sun, Luca Weihs, Eli VanderBilt, Alvaro Herrasti, Winson Han, Jiajun Wu, Nick Haber, Ranjay Krishna, Lingjie Liu, et~al.
\newblock Holodeck: Language guided generation of 3d embodied ai environments.
\newblock In \emph{2024 IEEE/CVF Conference on Computer Vision and Pattern Recognition (CVPR)}, pp.\  16277--16287. IEEE, 2024.

\bibitem[Yao et~al.(2025)Yao, Zhang, Yan, Zeng, Zhang, Xu, Yang, Gu, and Yu]{yao2025cast}
Kaixin Yao, Longwen Zhang, Xinhao Yan, Yan Zeng, Qixuan Zhang, Lan Xu, Wei Yang, Jiayuan Gu, and Jingyi Yu.
\newblock Cast: Component-aligned 3d scene reconstruction from an rgb image.
\newblock \emph{ACM Transactions on Graphics (TOG)}, 44\penalty0 (4):\penalty0 1--19, 2025.

\bibitem[Zhang et~al.(2021)Zhang, Cui, Zhang, Zeng, Pollefeys, and Liu]{zhang2021holistic}
Cheng Zhang, Zhaopeng Cui, Yinda Zhang, Bing Zeng, Marc Pollefeys, and Shuaicheng Liu.
\newblock Holistic 3d scene understanding from a single image with implicit representation.
\newblock In \emph{2021 IEEE/CVF Conference on Computer Vision and Pattern Recognition (CVPR)}, pp.\  8829--8838. IEEE, 2021.

\bibitem[Zhang et~al.(2026)Zhang, Chen, Xu, Huang, Mei, Chen, Zhou, Yuan, Cai, Huang, et~al.]{zhang2026flatland}
Jiahui Zhang, Yurui Chen, Yueming Xu, Ze~Huang, Jilin Mei, Chunhui Chen, Yanpeng Zhou, Yu-Jie Yuan, Xinyue Cai, Guowei Huang, et~al.
\newblock From flatland to space: Teaching vision-language models to perceive and reason in 3d.
\newblock \emph{Advances in Neural Information Processing Systems}, 38, 2026.

\bibitem[Zhou et~al.(2024)Zhou, Liu, and Han]{zhou2024zero}
Junsheng Zhou, Yu-Shen Liu, and Zhizhong Han.
\newblock Zero-shot scene reconstruction from single images with deep prior assembly.
\newblock \emph{Advances in Neural Information Processing Systems}, 37:\penalty0 39104--39127, 2024.

\end{thebibliography}
